\documentclass[journal]{IEEEtran}
\ifCLASSINFOpdf
\else
\fi

\usepackage{multirow}
\usepackage{tikz}
\usepackage{adjustbox}

\begin{document}
\title{Face Morphing Attacks and Face Image Quality: The Effect of Morphing and the Unsupervised Attack Detection by Quality}

\author{Biying Fu,
        Naser Damer,~\IEEEmembership{Member,~IEEE,}
\thanks{B. Fu was with the Department
of Smart Living \& Biometric Technologies, Fraunhofer Institute for Computer Graphics Research, Darmstadt,
, 64283 Germany e-mail: biying.fu@igd.fraunhofer.de.}% <-this % stops a space
\thanks{N. Damer was with the Department of Computer Science, TU Darmstadt, Darmstadt, 64283 Germany.}% <-this % stops a space
}


\markboth{Journal of xxxxx, August~2022}%
{Fu \MakeLowercase{\textit{et al.}}: Face Morphing Attacks and Face Image Quality: The Effect of Morphing and the Unsupervised Attack Detection by Quality}



\maketitle

\begin{abstract}
Morphing attacks are a form of presentation attacks that gathered increasing attention in recent years. A morphed image can be successfully verified to multiple identities. This operation, therefore, poses serious security issues related to the ability of a travel or identity document to be verified to belong to multiple persons. 
Previous works touched on the issue of the quality of morphing attack images, however, with the main goal of quantitatively proofing the realistic appearance of the produced morphing attacks.
We theorize that the morphing processes might have an effect on both, the perceptual image quality and the image utility in face recognition (FR) when compared to bona fide samples.
Towards investigating this theory, this work provides an extensive analysis of the effect of morphing on face image quality, including both general image quality measures and face image utility measures.
This analysis is not limited to a single morphing technique, but rather looks at six different morphing techniques and five different data sources using ten different quality measures.
This analysis reveals consistent separability between the quality scores of morphing attack and bona fide samples measured by certain quality measures.
Our study goes further to build on this effect and investigate the possibility of performing unsupervised morphing attack detection (MAD) based on quality scores.
Our study looks intointra and inter-dataset detectability to evaluate the generalizability of such a detection concept on different morphing techniques and bona fide sources.
Our final results point out that a set of quality measures, such as MagFace and CNNNIQA, can be used to perform unsupervised and generalized MAD with a correct classification accuracy of over 70\%.
\end{abstract}


% \fi
\IEEEpeerreviewmaketitle

\section{Introduction}
The advances in the accuracy of FR, driven by innovative training strategies \cite{deng2019arcface,DBLP:journals/corr/abs-2109-09416} and network architectures \cite{DBLP:journals/access/BoutrosSKDKK22,DBLP:conf/icb/BoutrosDFKK21}, are making FR a method of choice for physical and logical access control. However, FR is still vulnerable to attacks such as the face morphing attack.  
Face morphing incorporates two or more faces from different individuals to create a new face image, such that the newly created face image (the morphed image) can be successfully verified to multiple identities \cite{DBLP:conf/icb/FerraraFM14}. Therefore this operation poses a high potential risk in areas such as border control or financial transactions. 
Typically, a morphed face image does not only has to be verifiable to multiple identities but also has to appear realistic in case of human inspection \cite{DBLP:conf/btas/DamerBSKK19, DBLP:journals/corr/abs-2009-01729}.
Therefore, different works presented new morphing methodologies that focus on image appearance \cite{ReGenMorph,DBLP:conf/btas/DamerBSKK19, DBLP:journals/corr/abs-2009-01729, DBLP:journals/iet-bmt/NeubertMHKD18}. 

As the morphing process can leave morphing artifacts, whether performed on the image level or on a representation level in a generative framework, these artifacts can theoretically effect the perceptual quality of the image or even the utility of the image for face recognition.
Few previous works looked at some aspects of the morphed image quality, especially in relation to the bona fide images, but with very limited quality metrics and for the sole reason of proving that the presented morphing processes produce images that are similar to the bona fide ones \cite{DBLP:journals/corr/abs-2009-01729,DBLP:conf/btas/DamerBSKK19,DBLP:conf/iciap/DebiasiDSRSBKU19}.
Our recent work \cite{DBLP:conf/biosig/FuSCD21}, which this manuscript extends on, went into more details in measuring the effect of morphing on face image quality (FIQ). The work pointed out that the morphing process did result in a consistent effect on certain quality measures, even when re-digitization is applied to the image.
However, this initial work was limited by investigating the effect on only a single landmark-based morphing technique, namely the OpenCV-based morphing \cite{Satya}. The work did not also consider the possibility of detecting morphing attacks based on quality.

Following the highly interesting outcome of \cite{DBLP:conf/biosig/FuSCD21}, but the limited generalizability and practicality of its final conclusions, this paper is presented as an extended version of the work done in \cite{DBLP:conf/biosig/FuSCD21}, which achieved the best poster award at BIOSIG 2021. In this work, we extend the previous contributions as follows:
\begin{itemize}
    \item Instead of a single morphing technique, this work extends the study of the effect of morphing on face image quality to six different morphing techniques.
    \item Instead of a single source of bona fide and morphed face images, this work extends that to five diverse morphing datasets, each with multiple variations in the creation or post-processing of the attacks.
    \item A major extension is the novel exploration of using the quality scores as an unsupervised tool for the detection of face morphing attack and its generalizability across datasets and attack types.
\end{itemize}

Towards the added value of this extended study, the two main contributions of this work are:
\begin{itemize}
    \item Theorize a link between face morphing and face image quality and utility. Based on that, we uncover and analyse the effect of face morphing on face image quality and utility and the stability of this effect across different morphing methods.
    \item Propose the use of face image quality and utility measures as an unsupervised MAD method by taking advantage of the morphing process effect on the quality measures. We additionally investigate the generalizability of this proposed MAD concept across various variations that might face MAD in real applications.
\end{itemize}

These contributions thus drive two basic research questions addressed in this work: 1) what is the effect of morphing on face image quality and utility? and 2) can this effect be leveraged as an indication to perform MAD in an unsupervised manner? 
To address these questions and reach to a clear confirmation on our contributions, we perform experiments on 10 quality measures, divided into face image quality assessment (FIQA) and general image quality assessment (IQA) categories, experimented on 5 morphing datasets that include six different morphing techniques and five different data sources. Our analyses towards answering the main research questions involved the following main detailed investigations:
\begin{itemize}
    \item Studying the effect of a wide set of morphing techniques on the image qualities, measured by a diverse set of quality measures, in comparison to different bona fide sources.
    \item Revealing the different levels of separability between quality scores produced by different quality measures on bona fide vs. morphing attacks. 
    \item Investigating the quality differences between different bona fide sources when the quality is measured by different quality measures.
    \item Probing the possibility of leveraging the quality effect to detect face morphing attacks (i.e. classifying bona fide and morphing attacks) by using the quality scores as an MAD decision score within each morphing dataset.
    \item Investigating the generalizability of classifying bona fide samples of different sources when the quality scores are used as an MAD decision score and decision thresholds are set on other unknown bona fide sources.
    \item Investigating the generalizability of detecting morphing attacks from different morphing techniques and unknown data sources when the decision thresholds are set on other setup containing unknown morphing attacks and bona fide sources.
    \item Revealing the overall possible performance and generalizability of MAD based on different quality metrics on a wide set of attacks and data sources.
\end{itemize}
    
Our study points out that for certain quality and utility metrics, such as CNNIQA \cite{kang2014convolutional} and MagFace \cite{meng2021magface}, there is a significant effect of morphing on the estimated qualities. Such methods also produced similar quality values for different sources of bona fide images.
Using selected quality measures can lead to overall MAD accuracy above 70\%  on a diverse set of unknown attack methods and data sources, without the explicit MAD training.

We structured our work in the following by first introducing relevant works in Section \ref{ch:related} with respect to works relating image quality to morphing attacks, where the quality of morphs are however mostly compared to bona fide source images and the detection of morphing attacks. In Section \ref{ch:quality_metrics}, we describe the quality and utility measures studied in this work to investigate how the morphed images effect the image quality and if these findings can be leveraged on the MAD task. The morphing datasets and their used morphing techniques are presented in Section \ref{ch:experiments} followed by the experimental design with the used evaluation metrics and the investigation protocols. Finally, the results are thoroughly discussed in Section \ref{ch:discussion} and concluded in the Section \ref{ch:conclusion}.

\subsection{Nomenclature}
To enable a clear understanding of the experimental design and discussion, we introduce here a list of the used terminologies that are essential to be differentiated.
\begin{itemize}
    \item A morphing "dataset" refers to a complete set of data that includes both, the morphing attacks and the bona fide images. The used morphing datasets are detailed in Section \ref{sec:datasets}.
    \item A "data source" is the face image dataset where the bona fide samples (and the samples used to create the morphing attacks) of each morphing dataset originate.
    \item A "morphing technique" is a process that creates a morphing attack from a two or more bona fide samples. The morphing methods considered in this work are presented in details in Section \ref{sec:morphing_techniques}.
    \item A "morphing attack" is an image resulting from morphing images belonging to multiple identities and aim at being positively verifiable these identities.
    \item An "image quality assessment method (IQA)" is a method that assess the perceived or statistical quality of an image and not specifically designed to measure the utility of the image for face recognition.
    \item A "face image quality assessment method (FIQA)" is a method that assess the utility of a face image for face recognition.
    \item An "unknown morphing method" is a morphing method not used in the training of a specific MAD, and thus "unknown" to the MAD approach.
    \item An "unknown dataset" is a morphing dataset that is not used in the training of a specific MAD, and thus "unknown" to the MAD approach.
    \item An "unknown data source" is a data source that is not used in the training of a specific MAD, and thus "unknown" to the MAD approach.
    \item The "intra-dataset detectability" is the experimental setup where the performance of MADs is measured on the same morph dataset used for training, although on an identity-disjoint test set not used for training.
    \item The "inter-dataset detectability" is the experimental setups where the performance of MADs is measured on morphing datasets not used for training the MADs. 
\end{itemize}

\section{Related Works}\label{ch:related}

Face recognition systems (FRS) are highly vulnerable to face morphing attacks \cite{ferrara2014magic}. This poses a severe security risk to operations using automatic face recognition systems for authentication. Therefore, it is of great interest to understand the effect of such attacks and to have effective defense mechanisms. This section of related work thus gives a brief but concise view on two aspects, the previous works touching on the relationship between morphing and image quality, and a high-level view on MAD.

\subsection{Morphing attacks and quality}

In \cite{DBLP:conf/iciap/DebiasiDSRSBKU19}, Debiasi et al. investigated the qualities on morphing attacks, by using image quality (IQ) measures like BIQI \cite{DBLP:journals/spl/MoorthyB10}, BRISQUE \cite{mittal2012no}, OG-IQA\cite{DBLP:journals/spic/LiuHZHB16} and SSEQ \cite{DBLP:journals/spic/LiuLHB14}. These methods are all based on studying the general image qualities and belong to the group of no-reference blind IQA methods, which require no bona fide source reference image as a comparison to assess the morphing attack's image quality. Debiasi uses these quality measures to investigate the vulnerability of the morphing attacks on the face recognition solutions. The intuition behind this is based on Ferrara et al.'s work \cite{ferrara2014magic} pointing out that human experts have difficulties recognizing morphed face images of high quality and bona fide images. Therefore, the authors herein \cite{DBLP:conf/iciap/DebiasiDSRSBKU19} believed that the quality of morphs might be linked to image quality. The experiments are conducted on the MorGAN dataset which contains the bona fide images, MorGAN, and OpenCV generated morphs. Results show similar quality distributions of MorGAN attacks to the bona fide images within the MorGAN dataset, while landmark-based attacks are slightly off in terms of image quality. This makes the detectability of these GAN-based attacks more difficult compared to classical approaches. 

Regarding generating morphs with high quality and resolution, to make the morphing attack more effective. The work in \cite{DBLP:conf/btas/DamerBSKK19} proposed an approach to enhance the GANs generated images into a more realistic quality and resolution morphing attacks. This improvement could further suppress the artifacts, but not effect their performance on the face recognition system. To evaluate the quality of the generated attacks, single quality measures are used as suggested in \cite{wasnik2017assessing}.
The authors reported multiple quality factors, such as sharpness \cite{gao2007standardization}, blur \cite{narvekar2011no}, exposure \cite{shirvaikar2004optimal}, global contrast factor (GCF)
\cite{matkovic2005global}, contrast \cite{zohra2017linear, gao2007standardization}, and brightness \cite{zohra2017linear}. These quality measures are reported both for bona fide and morphing attacks. The intuitions behind these quality factors are, the closer the measure to the bona fide quality factor is, the better is the generated morph image. Based on these outcomes, \cite{DBLP:conf/btas/DamerBSKK19} showed that the images enhanced with the proposed method are of higher visual quality compared to unprocessed attacks and attacks enhanced by other state-of-the-art super-resolution methods. 

Zhang et al.\cite{DBLP:journals/corr/abs-2009-01729} proposed a new morph generation approach using an Identity Prior Driven Generative Adversarial Network (MIPGAN) to create high quality and high resolution morphed facial images with minimal artifacts. This is enabled by using a specific loss function enforcing high perceptual quality and identity factors to keep generated morphs without loss of identity. The MIPGAN Face Morph Dataset is created from FRGC-V2 face dataset using only high-quality passport-like face images from 140 data subjects. 
The analysis of the perceptual image quality of morphs is performed by comparing it to the reference image. This is contradictory to the other previously mentioned quality measures, which favored the no-referenced quality assessment methods. The peak signal-to-noise ratio (PSNR \cite{turaga2004no}) and the structural similarity index measure (SSIM \cite{dosselmann2011comprehensive}) are used to evaluate the quality of the generated attacks. Results based on these quality measures showed (1) little deviation in the perceived quality for all four different types of morphing attacks and (2) the MIPGAN created morphs have slightly higher image quality compared to StyleGAN 2, and (3) MIPGAN morphs present similar quality to facial landmarks-based morphing attacks. In addition, the authors asked 56 human observers to make a visual comparison between pairs of images to assess the morphing qualities besides the quantitative measures like PSNR and SSIM. They found out that the detection of morphed face images is a challenging task even for human observers, both the experienced and inexperienced user groups when showing only one single image at a time. 

The recent work \cite{DBLP:conf/biosig/FuSCD21} investigated in total 12 quality measures on the LMA-DRD morphing dataset showing the effect of the morphs on quality and utility. The morphing dataset contains only one morphing technique based on OpenCV. To measure the face utility, four FIQA methods (MagFace\cite{meng2021magface}, SER-FIQ \cite{terhorst2020face}, FaceQnet \cite{hernandez2019faceqnet}, and rankIQ \cite{chen2015face}) are utilized to determine a quality value which is directly related to the term face utility for a FR algorithm. Overall eight general IQA methods are used to assess the perceived image quality on morphing attacks. These IQA methods include BRISQUE \cite{mittal2012no}, PIQE \cite{venkatanath2015blind}, NIQE \cite{mittal2012making}, CNNIQA \cite{kang2014convolutional} , DeepIQA \cite{bosse2017deep}, MEON \cite{ma2017end}, rankIQA \cite{liu2017rankiqa}, and dipIQ \cite{7934456}. All selected quality and utility measures do not require a bona fide source to evaluate the attack's quality. 
Their results stated that most investigated IQ measures show only minor shift in quality for bona fide images compared to morphing attacks. Only MagFace \cite{meng2021magface} from the FIQA methods shows a clear separability between morphing attacks and bona fide images. Less differentiation between the re-digitization process and the digital image is detected by MagFace in terms of separability, which points out that the difference is correlated to the morphing process itself, while most IQA solutions do differentiate between the digital and re-digitized images but less on morphing attack itself. However, the results are limited to a single morphing technique and bona fide source. No studies on the detectability of morphing attack based on these quality and utility measures was conducted.

While most previous works used quality measures to support their claim of producing attacks similar to bona fide images, the contributions in \cite{DBLP:conf/biosig/FuSCD21} that we substantially extend here was limited to studying the effect on one morphing technique and one data source, without considering the generalizability of this effect, nor leveraging this effect to be used as an unsupervised MAD methods. This motivates our work to present novel contributions in these regards.

\subsection{Morphing attacks detection}

MAD methods can be categorized into two main groups, single image and differential MAD \cite{survey}. Single image MAD only analyses the investigated image to make a predicted decision of attack or bona fide \cite{DBLP:conf/icb/RaghavendraRVB17,DBLP:conf/cvpr/RaghavendraRVB17a,DBLP:conf/fusion/DamerZWSKK19,DBLP:conf/iwbf/AghdaieCSDN21,DBLP:journals/corr/abs-2103-01924}. Differential MAD analyses an investigated image along with a live image (assuming that the process allows for that). Differential MAD analyses the relation between both images to build a decision of attack or bona fide  \cite{DBLP:conf/dagm/DamerBWBTBK18,DBLP:journals/tifs/ScherhagRMB20,DBLP:conf/icb/DamerSZWTKK19,DBLP:conf/icpr/SoleymaniCDDN20}. However, the applicability of differential MAD is limited by the requirement of a live image and thus might be less practical in various use-cases.

Single image MAD solutions are commonly modeled as a binary classifier and can be roughly categorized into ones using handcrafted features and ones using deep learning features. 
Such handcrafted features included Binarized Statistical Image Features (BSIF) \cite{DBLP:conf/iwbf/ScherhagRRGRB17,DBLP:conf/btas/RaghavendraRB16}, Local Binary Patterns (LBP) \cite{DBLP:conf/btas/DamerS0K18}, Local Phase Quantization (LPQ) \cite{DBLP:conf/icb/RaghavendraRVB17}, or features established in the image forensic analyses such as the photo response non-uniformity (PRNU) \cite{DBLP:conf/iciap/DebiasiDSRSBKU19}. The MAD solutions based on deep learning commonly used pre-trained networks with or without fine-tuning, such as versions of VGG \cite{MADVgg}, AlexNet \cite{DBLP:conf/cvpr/RaghavendraRVB17a}, or networks trained for face recognition purposes such as OpenFace \cite{DBLP:conf/fusion/DamerZWSKK19}. However, all these works used a single binary label as the target of their training. A recent work leveraged the use of pixel-wise supervision towards creating a more generalizable MAD \cite{DBLP:journals/corr/abs-2103-01924}. The work showed higher generalizability over re-digitized samples when using the proposed training paradigm. The use of synthetic-based face images to create a diverse MAD development data was recently proposed and proved to enhance the performance and generalizability of MADs in comparison to MADs developed based on authentic data \cite{DBLP:journals/corr/abs-2203-06691}, especially given that the authentic data is constrained by its lack of diversity and the ethical/privacy issues related to using and sharing it.

A number of MAD works have analysed the issue of the generalizability of the MAD decisions when facing variabilities in the face morphing or image handling process. Such variabilities included the synthetic image generation processes \cite{DBLP:conf/btas/DamerGZKK19,DBLP:conf/btas/DamerBSKK19,DBLP:journals/corr/abs-2009-01729,ReGenMorph}, different data sources \cite{DBLP:conf/iwbf/ScherhagRB18}, morphing pair selection \cite{DBLP:conf/icb/DamerSZWTKK19},
image compression
\cite{DBLP:conf/visapp/MakrushinND17}, and re-digitization  \cite{MADVgg,DBLP:conf/btas/RaghavendraRB16,DBLP:conf/cvpr/RaghavendraRVB17a}. These variabilities have been shown to cause a drop in the MAD performance when they were unknown in the MAD training phase. A research direction towards avoiding the dependency on training data, and thus enhance the generalizability in detecting attacks with unknown variabilities, is the unsupervised detection of attacks as anomalies. MAD based on anomaly detection was only rarely addressed in the literature \cite{DBLP:conf/btas/DamerGZKK19} based on one-class classification of handcrafted and deep learned features, however, with very limited detection performance \cite{DBLP:conf/btas/DamerGZKK19}. This work, in its effort to investigate the possibility of using FIQA and IQA as indicators of an image being morphed, also offers an unsupervised MAD method that operates without the need to train an MAD to detect already known attacks. 

\section{Face image quality and general image quality }\label{ch:quality_metrics}

FIQ represents the utility (value) of the face image to FR algorithms \cite{ISOIEC2382-37,DBLP:journals/corr/abs-2112-06592}. This utility can be measured by the FIQ score (scalar) following the definition in ISO/IEC 2382-37 \cite{ISOIEC2382-37}. The algorithms designed to measure this utility metric belong to the family of FIQA methods. The general image quality, on the other hand, focuses on perceived image quality and it does not necessarily reflect the utility of the image in FR, e.g. an image of a mask occluded face can be of high perceived quality but of low utility to FR algorithms \cite{DBLP:conf/fgr/FuKD21}. 
To investigate the effect of face morphing on the image and the possibility to use this effect in detecting morphing attacks, we select a total of 10 different quality measures, divided between IQA and FIQA categories, which will be summarised in this section.

The morphing processes do introduce changes to the bona fide morphed images. These images can be, to variant degrees, apparent according to the morphing process, see Figures \ref{fig:samples_morph} and \ref{fig:samples_bf}.
We theorize that the artifacts, whether minor or major, might have an effect on the perceived image quality measured by IQA.
These artifacts, as well as the fact that the image becomes less distinct (can be matched to multiple identities), are also theorized to have an effect on the utility of the image in FR, which can be measured by FIQA.
This leads to the wide range of investigations in this work, addressing both, the effect of different morphing techniques on the IQ and FIQ of the images, as well as the possible use of this effect in detecting morphing attacks.

\subsection{General image quality assessment}

The five IQA approaches selected in this paper can be grouped into the following four sub-categories: (1) based on natural image statistics \cite{mittal2012no}, (2) convolutional neural network-based \cite{kang2014convolutional}, (3) multi-task learning-based approach \cite{DBLP:journals/tcsv/ZhangMYDW20}, and (4) ranking-based methods \cite{DBLP:journals/tip/ZhangMZY21,7934456}. These assessed qualities by these methods are typically affected by image distortions, artifacts, and perceived degradation. 

\textbf{BRISQUE} \cite{mittal2012no} accounts for the first sub-category and is learned based on studying the deviation from the general statistics of natural images. These statistics are originated from the finding by Rudermann \cite{ruderman1994statistics} that assumes natural scene images have a distribution similar to a normal Gaussian distribution. The degree of deviation from the normal Gaussian is thus directly related to the degradation in image quality. 

The second sub-category uses convolution layers to automatically extract deep features without the need for a priori special design using handcrafted features. \textbf{CNNIQA} \cite{kang2014convolutional} is accounted to this category. This approach is a patch-based approach that correlates the image quality based on the fusion of patch decisions, enabling local decisions to communicate to form a global decision. The default setting with a patch size of 32x32 and a kernel size of 7x7 is chosen according to the original publication in \cite{kang2014convolutional}. However, we applied the sliding window approach with a stride size of 4 pixels to increase the patch resolution as the input image has only a reduced image resolution compared to the training images used in \cite{kang2014convolutional}.

\textbf{DBCNN}~\cite{DBLP:journals/tcsv/ZhangMYDW20} is another general IQA method based on Multi-task learning. In the core of the algorithm, two CNNs were trained, one on large-scale synthetically generated datasets, while the other focuses on the classification network pre-trained on ImageNet \cite{DBLP:conf/cvpr/DengDSLL009} for extracting more authentic distortions. The final quality score is pooled from the features of both CNNs to a unified representation.

The methods \textbf{UNIQUE}~\cite{DBLP:journals/tip/ZhangMZY21} and \textbf{dipIQ}\cite{7934456} are categorized to ranking-based IQA approaches. One benefit of these methods is that it is easier to generate ranked image pairs since the absolute quality score for each image is hard to acquire. In addition, it is simple to introduce additional image distortions and synthesize ranked image pairs. Therefore the cost of generating a large, synthetic dataset with ranked image pairs is relatively low. The network structures of these methods are built using two parallel streams with shared weights. Only one trained stream is used to assess the quality of the input image in the prediction stage. 

\subsection{Face image quality assessment}

Recent FIQA methods take advantage of deep-learning-based approaches and a large amount of publicly available face image datasets. Thus, these methods show relatively good performance in predicting the utility of a face image for FR. The five selected FIQA methods are to be categorized into training with (1) supervised regression learning based on pseudo quality labels as in \cite{meng2021magface, terhorst2020ser}, (2) unsupervised methods based on FR model behavior as in \cite{hernandez2019faceqnet, DBLP:journals/corr/abs-2103-05977}, or (3) ranking-based approaches as in \cite{chen2015face}. 

We used two unsupervised FIQA methods, namely \textbf{MagFace} and \textbf{SER-FIQ} proposed by Meng et al. in \cite{meng2021magface} and Terhoerst et al. in \cite{terhorst2020ser}. In MagFace, the FIQ estimation is considered by the loss function during the training process. The face embedding learned by the network can be used both for FR and FIQ estimation. The magnitude of the unnormalized face embedding is proportional to the cosine distance to its class center and is also directly related to the face utility. In SER-FIQ, this method mitigates the need for any automated or human labeling by correlating the FIQ to the robustness of the extracted face embeddings to random dropout patterns. Face images with high utility are expected to have similar face representations resulting in low variance and vice versa. 

\textbf{FaceQnet}~\cite{hernandez2019faceqnet} and \textbf{SDD-FIQA}~\cite{DBLP:journals/corr/abs-2103-05977} are categorized into the supervised FIQA methods. While FaceQnet uses ICAO compliant face images as high-quality reference images and builds quality labeling scores based on the FR genuine comparison score to ICAO compliant images, the SDD-FIQA uses pseudo quality label based on the Wasserstein distance between similarity distributions of genuine and imposter image pairs. Both training networks then fine-tuned a pre-trained FR base-network (RseNet-50 \cite{he2016deep}) as the backbone architecture and the successive regression layer on top of the feature extraction layers are used to associate an input image to an FIQ score.

A ranking-based FIQA method learned on comparing pairs of input images is the \textbf{rankIQ}~\cite{chen2015face}. Chen et al. \cite{chen2015face} are inspired by the idea that it is easier for a human operator to rank pairs in a relative manner than to define an absolute quality score for a single image. This method combines two-stage learning to first extract individual handcrafted features (e.g., HoG, Gabor, LBP, and CNN features) and later map these feature scores to a final quality score by using a kernel trick. Compared to the other considered FIQA methods, this method does not relate to very deep features or deep backbones, however, proved to perform competitively to recent FIQA approaches \cite{DBLP:conf/wacv/FuCHD22}.

The presented methods in this section are designed for either assessing the general image quality or the face image utility (for FR) in the design concept. A number of the later methods were shown to predict low qualities for images where FR models focus on areas beyond the central face area \cite{DBLP:conf/wacv/FuD22}, which might point out extreme poses among other factors. Our work investigates the face morphing effect on these measures. Table \ref{tab:quality_metrics} listed the interpretation of these quality measures. The term "increasing" indicates the ascending order of this quality score and means the higher the score the better is the perceived quality or utility, and vice versa. 

\begin{table*}
\caption{The table shows the ordering of the quality measures considered in the manuscript. The term "increasing" means that the way this quality measure is ordered is ascending and the higher the quality score, the better the perceived quality or utility and vice versa. \label{tab:quality_metrics}}
{
\resizebox{\textwidth}{!}{
\begin{tabular}{lllll|lllll}
\hline
\multicolumn{5}{c|}{FIQA}                             & \multicolumn{5}{c}{IQA}                              \\ \hline
MagFace \cite{meng2021magface} & SDD-FIQA \cite{DBLP:journals/corr/abs-2103-05977} & SER-FIQ \cite{terhorst2020ser}  & FaceQnet v2 \cite{hernandez2019faceqnet} & rankIQ \cite{chen2015face}  & BRISQUE \cite{mittal2012no}  & DBCNN \cite{DBLP:journals/tcsv/ZhangMYDW20} & CNNIQA \cite{kang2014convolutional}   & UNIQUE \cite{DBLP:journals/tip/ZhangMZY21}  & dipIQ \cite{7934456}    \\ \hline
Increase & Increase & Increase & Increase & Increase & Decrease & Increase & Decrease & Increase & Increase \\ \hline
\end{tabular}
}
}{}
\end{table*}

\section{Experimental Design}
\label{ch:experiments}

The experiments in this work are designed so that they provide comprehensive answers to the following two research questions: (1) what is the effect of different morphing approaches on different IQA and FIQA measures, and (2) is it possible to use this effect to detect morphing attacks without the explicit training of an MAD.
To address these two main investigations, we structured our experiment section first to introduce the variety of different morphing techniques and the 5 morphing datasets studied in this work. Each morph dataset intends to provide variations in the created morphing attacks depending on the bona fide source distributions extracted from the respective source face datasets. We subsequently introduce the evaluation metrics before presenting the protocols used in our investigations.

\subsection{Morphing Techniques}
\label{sec:morphing_techniques}
The morphing techniques considered in this work can be grouped into classical landmark-based approaches and deep-learning-based approaches using generative adversarial networks (GANs). While the landmark-based morphing approaches could introduce visible blurring effects or other perceptual artifacts, the GAN-based approaches can generate more realistic morphing attacks. However, besides the difficult training procedure of GAN-based approaches, it is also more difficult to constrain the output of the GAN-based attacks. In the following, we will shortly explain these morphing techniques in their individual category.

\subsubsection{Facial Landmarks-Based Morphing Techniques}

The \textbf{OpenCV} algorithm is an open-source implementation proposed by Satya et al. in \cite{Satya} using the extracted locations of 68 facial landmarks determined by the Dlib library \cite{DBLP:journals/jmlr/King09}. The facial landmarks are used to transform both bona fide source images into one morphed image using Delaunay triangles to warp the facial landmarks and alpha blending. This morphing approach has been proven to produce attacks leading to a relatively high vulnerability in FR systems \cite{DBLP:journals/corr/abs-2009-01729, DBLP:journals/corr/abs-2103-01924, DBLP:journals/corr/abs-2012-05344}.  

\textbf{FaceMorpher} \cite{Alyssa} is another open-source and facial landmark-based morphing algorithm, which is similar to the OpenCV algorithm. Instead of using Dlib, this method is based on STASM \cite{DBLP:conf/visapp/MilborrowN14} as the facial landmark detector. These two landmarked-based morphing algorithms create morphed face images with visible artifacts especially around the hair region because the regions outside the landmark positions are simply averaged. 

\textbf{WebMorph} \cite{DeBruine} is an online tool provided by the FRLL dataset provider to generate morphed images. This method leveraged 189 facial landmarks that are specifically labeled for the FRLL dataset. The high amount of precisely labeled landmark positions can generate morphs with reduced visible artifacts compared to using only 68 facial landmarks. However, as this method works exclusively for annotated FRLL images, this morphing technique is used only on the FRLL dataset. 

\textbf{AMSL} is introduced by Neubert et al. in \cite{DBLP:journals/iet-bmt/NeubertMHKD18} generating morphed images from bona fide images of FRLL dataset \cite{DeBruine_FRLL} using a private Combined Morphs tool. This tool enables generating morphed images with no visible ghosting artifacts around the hair and neck areas compared to OpenCV, FaceMorpher, and WebMorph, due to its post-processing step like Poisson image editing. 

\subsubsection{Generative Adversarial Network-based Morphs}

Leveraging the recent advances in deep-learning and generative adversarial networks (GANs), these GAN-based methods can generate more photo-realistic images without visible artifacts such as those blending artifacts introduced from other landmark-based morphing techniques. \textbf{StyleGAN 2} \cite{DBLP:conf/cvpr/KarrasLAHLA20} is pre-trained on FFHQ dataset introduced in \cite{DBLP:conf/cvpr/KarrasLA19} and can generate high resolution and realistic faces without introducing noticeable artifacts. The StyleGAN was first used in \cite{DBLP:conf/iwbf/VenkateshZRRDB20} to produce morphed images, which later used an updated generator in \cite{DBLP:journals/corr/abs-2009-01729}.

\textbf{MorGAN} \cite{DBLP:conf/btas/DamerS0K18} was the first proposed generative approach. MorGAN uses a specific loss function so that instead of generating a morph with most probable lost identity using the GAN training, the network enforces the generated morphs to keep the identity information. However, the output of the MorGAN generated morphs are relatively unrealistic with a resolution of 64x64 pixels.

\subsection{Datasets}
\label{sec:datasets}
Five different and diverse morphing datasets are investigated in this work to provide generalized conclusions. They compose of FRLL-Morphs \cite{DBLP:journals/corr/abs-2012-05344, DeBruine_FRLL}, FERET-Morphs \cite{DBLP:journals/corr/abs-2012-05344, PHILLIPS1998295}, and FRGC-Morphs \cite{DBLP:journals/corr/abs-2012-05344, DBLP:conf/cvpr/PhillipsFSBCHMMW05}, LMA-DRD \cite{DBLP:journals/corr/abs-2103-01924}, and MorGAN \cite{DBLP:conf/btas/DamerS0K18}.

Research in the field of face MAD becomes more stringent, but there is a lack of shared, publicly available datasets. To overcome this gap, Sarkar et al. \cite{DBLP:journals/corr/abs-2012-05344} worked on creating publicly available morphing datasets with several forms of attacks to be used for research purposes. They provided new morphing datasets with five different types of morphing attacks based on OpenCV  \cite{Satya}, FaceMorpher \cite{Alyssa}, WebMorph  \cite{DeBruine}, AMSL \cite{DBLP:journals/iet-bmt/NeubertMHKD18} and StyleGAN 2 \cite{DBLP:conf/cvpr/KarrasLAHLA20}. The attacks are created using source face images from three publicly available face datasets including the Face Research London Lab (FRLL) dataset, the Facial Recognition Technology (FERET) dataset, and the Face Recognition Grand Challenge (FRGC) dataset. These three morph datasets also build a big part of the foundation of our research in this work. 

The \textbf{FRLL-Morphs} dataset \cite{DBLP:journals/corr/abs-2012-05344} is generated from the publicly available Face Research London Lab dataset \cite{DeBruine_FRLL}. This dataset include 5 different morphing techniques, including OpenCV \cite{Satya}, FaceMorpher \cite{Alyssa}, StyleGAN 2 \cite{DBLP:conf/cvpr/KarrasLAHLA20}, WebMorph \cite{DeBruine} and AMSL \cite{DBLP:journals/iet-bmt/NeubertMHKD18}. Each morph technique contains 1222 morphed face images generated using only frontal face images with high resolution. We note, however, that the AMSL generates two unique morphed images for every pair of source bona fide images, which doubles the size of the AMSL morphs compared to other morphing techniques. The bona fide face images are extracted using the provided protocol from the FRLL-Morphs dataset \cite{DeBruine_FRLL} using only the frontal and smiling face images. Even though the source images are of very high visual quality and under uniform illumination with a large variety of ethnicity, pose, and expression, the number of bona fide samples in the dataset is limited to 204 images from 102 identities.

The \textbf{FERET-Morphs} dataset \cite{DBLP:journals/corr/abs-2012-05344} and the \textbf{FRGC-Morphs} dataset \cite{DBLP:journals/corr/abs-2012-05344} are extracted from the official FERET \cite{PHILLIPS1998295} and FRCG v2.0 dataset \cite{DBLP:conf/cvpr/PhillipsFSBCHMMW05}, respectively. Both datasets include 3 different morphing techniques, including OpenCV, FaceMorpher, and StyleGAN 2. For each morphing technique, it contains 529 and 964 morphed face images for FERET-Morphs and FRGC-Morphs respectively. Opposite to FRLL, the FERET and FRCG v2.0 dataset contained face images of large variability in quality and thus enabling creating morphing attacks with larger quality distributions compared to FRLL-Morphs as can be seen in Figure \ref{fig:distributions}. The bona fide images are extracted using the provided protocols in the FERET-Morphs and FRGC-Morphs datasets.

THE \textbf{LMA-DRD} Morph dataset \cite{DBLP:journals/corr/abs-2103-01924} contains morphed face images created from the VGGFace2 \cite{DBLP:conf/fgr/CaoSXPZ18} dataset. The images used to generate morphs are frontal, with a neutral expression, and according to the generation details described in \cite{DBLP:journals/corr/abs-2103-01924}. The large VGGFace2 allows the selection of high-quality images according to ICAO standards. The morphing technique is based on OpenCV using facial landmarks. For further details of the morphing process and the parameters used, we refer to the work in \cite{DBLP:journals/corr/abs-2103-01924}. An additional bona fide image is selected for each morphed identity, when available. In total, the used LMA-DRD morphing dataset contains 276 digital Bona fide (D-BF) images and 364 digital morphing attacks (D-M). These images were printed on 11,5cmx9cm glossy photo paper in a professional studio and scanned with a 600dpi scanner. They resulted in the same number of re-digitized bona fide (PS-BF) and attacks (PS-M), leading to two versions of the dataset, the digital LMA-DRD (D) and the re-digitized LMA-DRD (PS). Opposite to other morphing datasets, LMA-DRD dataset specifically aims at investigating the morphing attacks after the re-digitization process. The re-digitized versions are intended to simulate real identity document issuing scenarios and their effect on the digital artifacts, which simulate a realistic challenge for MAD algorithms. 

To enable selection of similar faces to generate more sophisticated morphs, the \textbf{MorGAN} dataset \cite{DBLP:conf/btas/DamerS0K18} created morphing attacks from a large pool of identities using the CelebA \cite{DBLP:conf/iccv/LiuLWT15}. The CelebA dataset contains around 202,599 face images of 10,177 identities which allows the creation of morphing attacks fulfilling the intended goal. The MorGAN dataset contains 1000 morphed images for each of the two morphing attacks. One attack form is based on MorGAN and the other is created using the landmark-based morphing approach using OpenCV \cite{Satya}. The bona fide images contain two sets (1500 bona fide references and 1500 bona fide probes). The exact description of the morphing dataset can be found in \cite{DBLP:conf/btas/DamerS0K18}.  

In Table \ref{tab:dataset}, the presented 5 morphing datasets with their respective source dataset and the included morphing techniques are listed. In addition, the number of the used images in the individual setting is provided along with the bona fide image numbers and the total amount of images.  

\begin{table*}
\caption{The used morphing datasets are depicted in the table including their source dataset and the morphing techniques used in the individual dataset to create the morphing attacks. In addition, the number of the used images under each setting is also provided. \label{tab:dataset}}
\centering
{
\begin{tabular}{llllll}
\hline
                              &                          & \multicolumn{3}{c}{No of Images}                   &       \\ \hline
Morph dataset                & Source dataset        & no. bona fide         & morphing type & no. attack & Total \\ \hline
\multirow{5}{*}{FRLL-Morphs \cite{DBLP:journals/corr/abs-2012-05344}}  & \multirow{5}{*}{FRLL \cite{DeBruine_FRLL}}    & \multirow{5}{*}{204}  & AMSL \cite{DBLP:journals/iet-bmt/NeubertMHKD18}          & 2444       & 2648  \\
                              &                          &                       & FaceMorpher \cite{Alyssa}   & 1222       & 1426  \\
                              &                          &                       & OpenCV \cite{Satya}        & 1222       & 1426  \\
                              &                          &                       & StyleGAN 2  \cite{DBLP:conf/cvpr/KarrasLAHLA20}     & 1222       & 1426  \\
                              &                          &                       & WebMorph \cite{DeBruine}    & 1222       & 1426  \\ \hline
\multirow{3}{*}{FERET-Morphs \cite{DBLP:journals/corr/abs-2012-05344}} & \multirow{3}{*}{FERET \cite{PHILLIPS1998295}}   & \multirow{3}{*}{1413} & FaceMorpher \cite{Alyssa}   & 529        & 1942  \\
                              &                          &                       & OpenCV \cite{Satya}        & 529        & 1942  \\
                              &                          &                       & StyleGAN 2  \cite{DBLP:conf/cvpr/KarrasLAHLA20}      & 529        & 1942  \\ \hline
\multirow{3}{*}{FRGC-Morphs \cite{DBLP:journals/corr/abs-2012-05344}}  & \multirow{3}{*}{FRGC v2 \cite{DBLP:conf/cvpr/PhillipsFSBCHMMW05}} & \multirow{3}{*}{3167} & FaceMorpher \cite{Alyssa}   & 964        & 4131  \\
                              &                          &                       & OpenCV \cite{Satya}        & 964        & 4131  \\
                              &                          &                       & StyleGAN 2  \cite{DBLP:conf/cvpr/KarrasLAHLA20}     & 964        & 4131  \\ \hline
LMA-DRD (D) \cite{DBLP:journals/corr/abs-2103-01924}                  & VGGFace2 \cite{DBLP:conf/fgr/CaoSXPZ18}                & 276                   & OpenCV (D) \cite{Satya}    & 364        & 640   \\
LMA-DRD (PS) \cite{DBLP:journals/corr/abs-2103-01924}                & VGGFace2 \cite{DBLP:conf/fgr/CaoSXPZ18}                & 276                   & OpenCV (PS) \cite{Satya}   & 364        & 640   \\ \hline
\multirow{2}{*}{MorGAN \cite{DBLP:conf/btas/DamerS0K18}}       & \multirow{2}{*}{CelebA \cite{DBLP:conf/iccv/LiuLWT15}}  & \multirow{2}{*}{3000} & OpenCV \cite{Satya}       & 1000       & 4000  \\
                              &                          &                       & MorGAN \cite{DBLP:conf/btas/DamerS0K18}        & 1000       & 4000  \\ \hline
\end{tabular}

}{}
\end{table*}

\begin{figure*}
\centering
\begin{tabular}{ccccc}
    \includegraphics[width=.085\linewidth]{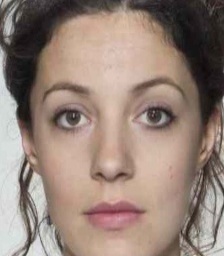}
    \includegraphics[width=.085\linewidth]{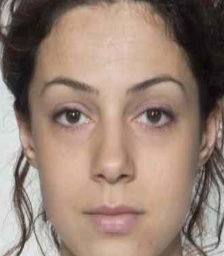} &  
    \includegraphics[width=.085\linewidth]{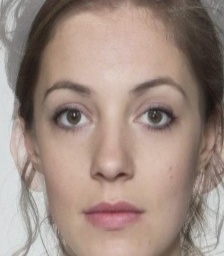}
    \includegraphics[width=.085\linewidth]{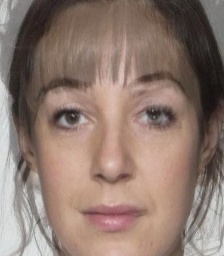} &  
    \includegraphics[width=.085\linewidth]{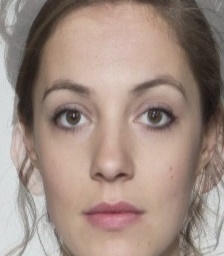}
    \includegraphics[width=.085\linewidth]{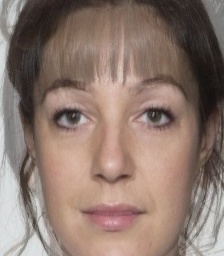} &  
    \includegraphics[width=.085\linewidth]{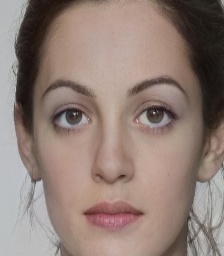}
    \includegraphics[width=.085\linewidth]{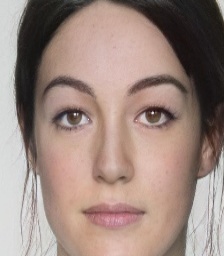} &      \includegraphics[width=.085\linewidth]{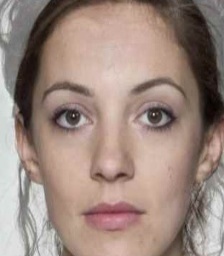}
    \includegraphics[width=.085\linewidth]{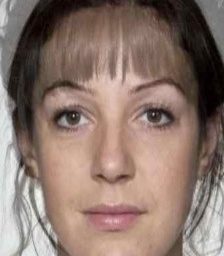}\\
    FRLL AMSL & FRLL FaceMorpher & FRLL OpenCV & FRLL StyleGAN 2 & FRLL WebMorph\\[6pt]
    
    \includegraphics[width=.085\linewidth]{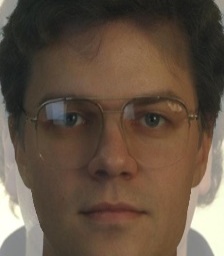}
    \includegraphics[width=.085\linewidth]{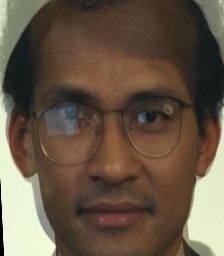} &  
    \includegraphics[width=.085\linewidth]{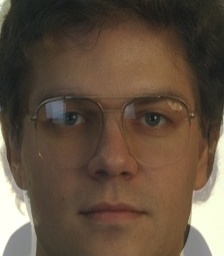}
    \includegraphics[width=.085\linewidth]{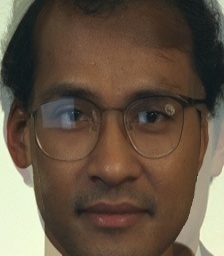} &  
    \includegraphics[width=.085\linewidth]{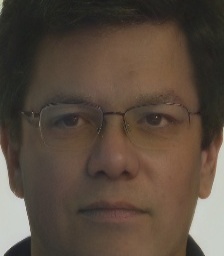}
    \includegraphics[width=.085\linewidth]{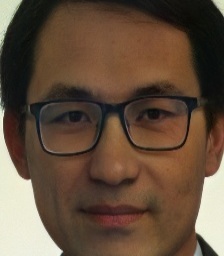} &  
    \includegraphics[width=.085\linewidth]{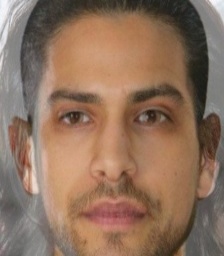}
    \includegraphics[width=.085\linewidth]{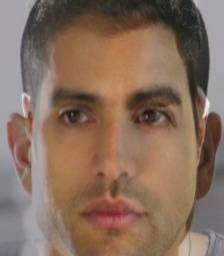} &        \includegraphics[width=.085\linewidth]{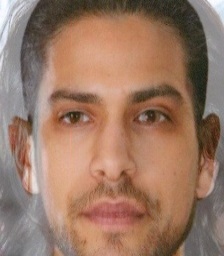}
    \includegraphics[width=.085\linewidth]{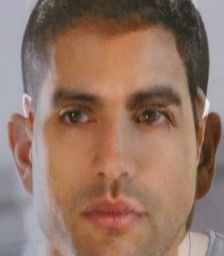}\\
    FERET FaceMorpher & FRLL OpenCV & FRLL StyleGAN 2& LMA-DRD (D) & LMA-DRD (PS) \\[6pt]
    
    \includegraphics[width=.085\linewidth]{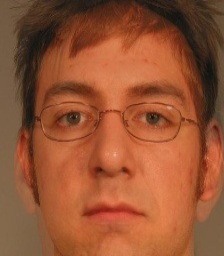}
    \includegraphics[width=.085\linewidth]{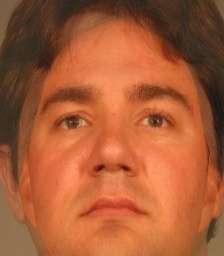} &  
    \includegraphics[width=.085\linewidth]{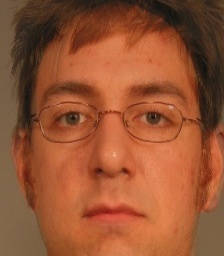}
    \includegraphics[width=.085\linewidth]{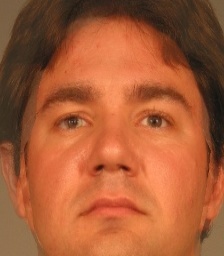} &  
    \includegraphics[width=.085\linewidth]{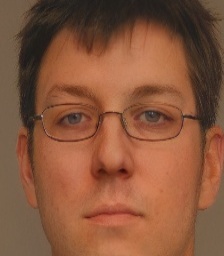}
    \includegraphics[width=.085\linewidth]{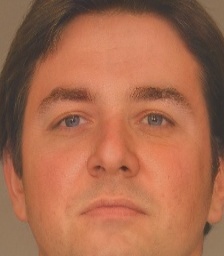} &  
    \includegraphics[width=.085\linewidth]{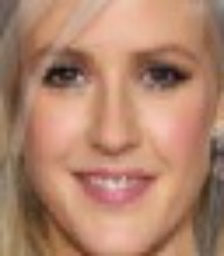}
    \includegraphics[width=.085\linewidth]{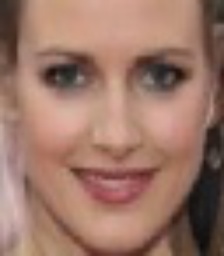} &        \includegraphics[width=.085\linewidth]{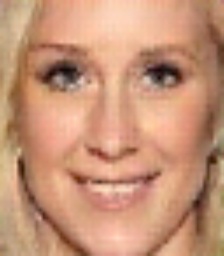}
    \includegraphics[width=.085\linewidth]{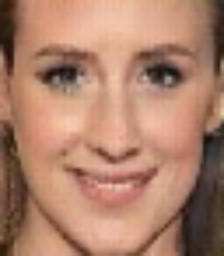}\\
    FRGC FaceMorpher & FRGC OpenCV & FRGC StyleGAN 2& MorGAN OpenCV & MorGAN MorGAN \\[6pt]
\end{tabular}
\caption{Each pair of images is an example of the morphed images generated by the morphing attack strategies from the considered 5 morphing datasets. The sub-caption first presents the name of the database followed by the naming of the morphing strategy. Visible artifacts in the morphed images are most apparent for landmark-based morphing techniques especially close to the nose and eyes region, while the GAN-based approaches generate morphed images with clearer background.}
\label{fig:samples_morph}
\end{figure*}

\begin{figure*}
\centering
\begin{tabular}{ccc}
    \includegraphics[width=.10\linewidth]{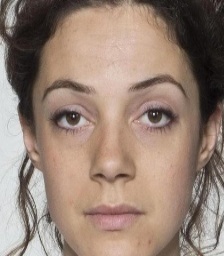}
    \includegraphics[width=.10\linewidth]{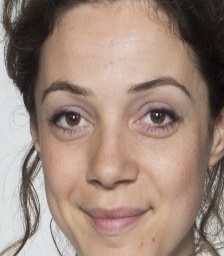} &  
    \includegraphics[width=.10\linewidth]{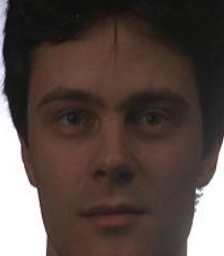}
    \includegraphics[width=.10\linewidth]{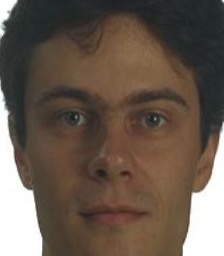} & 
    \includegraphics[width=.10\linewidth]{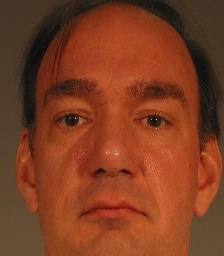}
    \includegraphics[width=.10\linewidth]{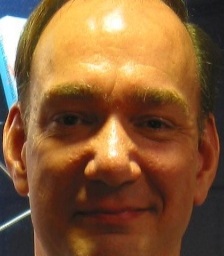} \\
    FRLL & FERET & FRGC \\[6pt]
    \includegraphics[width=.10\linewidth]{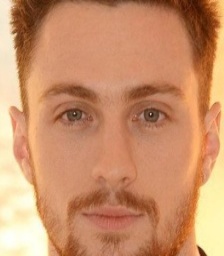}
    \includegraphics[width=.10\linewidth]{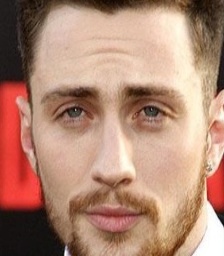} & 
    \includegraphics[width=.10\linewidth]{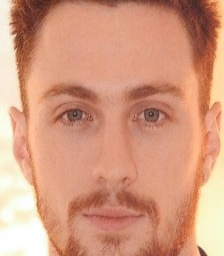}
    \includegraphics[width=.10\linewidth]{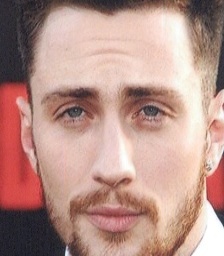} & 
    \includegraphics[width=.10\linewidth]{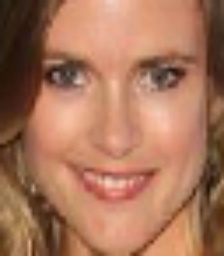}
    \includegraphics[width=.10\linewidth]{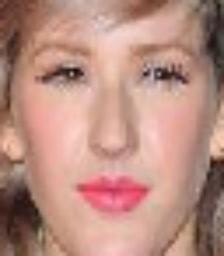}\\
    OpenCV (D) & OpenCV (PS) & MorGAN\\[6pt]
\end{tabular}
\caption{Each pair of images presents the bona fide images from the 5 morphing datasets.}
\label{fig:samples_bf}
\end{figure*}

For all images across the morphing datasets, the images are pre-processed using the MTCNN framework \cite{DBLP:journals/spl/ZhangZLQ16} to detect, crop, and align (geometric transform) the face to output size of 224x224 pixels whenever necessary. The face images in the MorGAN dataset are already pre-processed with an output size of 64x64 pixels. For each of the attack forms across the 5 morphing datasets (in total 15) a pair of example images are depicted in Figure \ref{fig:samples_morph}. Figure \ref{fig:samples_bf} further visualizes bona fide face image pairs for the 5 morphing datasets, where the LMA-DRD dataset has two different bona fides, one for the source digital images and the other for the re-digitized (PS) images. Table \ref{tab:samples} further listed the quality scores for the example images selected in both figures with 3 best performing quality measures from both the FIQA and IQA categories. For all quality measures, a higher quality score indicates a better quality of the face image, except for BRISQUE and CNNIQA, which have the inverted meaning. For most quality measures listed in the table, it is observed that the bona fide images have a higher quality compared to the morphing attacks across most morphing datasets.

\subsection{Performance Metrics}

As targeted to answer both (1) the effect of morphing on quality and (2) the detectability of morphing attacks by probing this effect, we grouped the used evaluation metrics into these two individual classes.

First, to consider the morphing attack and its effect on quality scores, we listed the quality scores for sample bona fide and attack images for reference. The different morphing dataset has different quality distribution. We show this by visualizing the quality score distribution approximated by the kernel density estimation method on the discrete quality scores of (1) the bona fide images across morphing datasets or (2) for bona fide images and the morphing attacks within the same morphing dataset. 

Another property to review is the separability between the quality distributions of morphing attacks and bona fide. Fisher Discriminant Ratio ($FDR$) is a useful measure to do just that. We looked at the quality score distributions of the individual attacks and the bona fide images within the same morphing dataset. The term FDR in \cite{damer2014biometric} is defined by Equation (\ref{eq:fdr}) and quantifies the separability of the score distributions of morphed and bona fide face images.
\begin{equation}
    FDR = \frac{(\mu_1 - \mu_2)^2}{(\sigma_1)^2 + (\sigma_2)^2},
    \label{eq:fdr}
\end{equation}
In Equation (\ref{eq:fdr}), the term $\mu_1$ and $\mu_2$ are the mean quality score of the morphed and bona fide distributions respectively and the $\sigma_1$ and $\sigma_2$ describe the standard deviation term of both distributions respectively.

\begin{figure*}[]
\centering
\foreach \db in {MagFace,SDD,SERFIQ,FaceQnet,rankIQ,BRISQUE,DBCNN,UNIQUE,CNNIQA,dipIQ}{
\foreach \model in {frll,feret,frgc,LMA-DRD_digital,LMA-DRD_ps,MorGAN}
{\includegraphics[width=0.167\linewidth]{\db_\model.jpg}}
}
\caption{Figure visualizes the distributions of the different morphs and the bona fide scores based on the investigated quality measures for the 5 morphing datasets, where the LMA-DRD further splits into the digital and re-digitized versions.}
\label{fig:distributions}
\end{figure*}

\begin{table*}
\caption{Quality scores of the displayed sample images from Figures \ref{fig:distributions} and \ref{fig:bona_fide_dist} are listed in here. A higher score indicates better quality except for BRISQUE and CNNIQA which possess a reversed ordering. This results in the observation of bona fide images having better quality compared to most of the morphing attacks across most morphing datasets, especially in the case of FRLL-Morphs, FERET-Morphs, and FRGC-Morphs. \label{tab:samples}}
{
\resizebox{\textwidth}{!}{
\begin{tabular}{llllllllllllll}
\hline
                               &                & \multicolumn{6}{c}{FIQA}                                                               & \multicolumn{6}{c}{IQA}                                                              \\ \hline
dataset                       & Morph Type     & \multicolumn{2}{c}{MagFace} & \multicolumn{2}{c}{SER-FIQ} & \multicolumn{2}{c}{rankIQ} & \multicolumn{2}{c}{BRISQUE} & \multicolumn{2}{c}{CNNIQA} & \multicolumn{2}{c}{dipIQ} \\ \hline
                               &                & left         & right        & left         & right        & left         & right       & left         & right        & left         & right       & left        & right       \\ \hline
\multirow{6}{*}{FRLL-Morphs}   & AMSL           & 26.9800      & 26.9126      & 0.8943       & 0.8928       & 68.6190      & 73.8720     & 12.9910      & 14.0530      & 23.4542      & 22.6041     & -3.1760     & -3.3660     \\
                               & FaceMorpher    & 27.1052      & 26.3664      & 0.8919       & 0.8889       & 71.9610      & 61.8510     & 10.3640      & 10.6400      & 19.1544      & 17.9703     & -0.4410     & -1.1200     \\
                               & OpenCV         & 27.3879      & 27.8763      & 0.8920       & 0.8928       & 74.0580      & 57.1580     & 8.8580       & 7.1440       & 18.4415      & 17.0790     & 0.0000      & -0.1540     \\
                               & StyleGAN 2       & 26.9321      & 27.5039      & 0.8916       & 0.8878       & 69.9110      & 70.2580     & 6.0070       & 8.8480       & 21.5867      & 19.5706     & -0.5050     & -1.8630     \\
                               & WebMorph     & 27.0006      & 27.5805      & 0.8912       & 0.8910       & 74.1640      & 59.6160     & 17.8020      & 14.3810      & 22.4474      & 19.8183     & -2.7940     & -4.4660     \\
                               & Bona Fide      & 28.4257      & 30.0034      & 0.8933       & 0.8930       & 74.5040      & 56.9300     & -0.1820      & 0.7070       & 20.3228      & 20.6871     & -2.7890     & -2.6990     \\ \hline
\multirow{4}{*}{FERET-Morphs}  & FaceMorpher    & 27.6738      & 26.1388      & 0.8907       & 0.8870       & 54.4640      & 65.0830     & 12.3540      & 9.0630       & 19.3835      & 18.1728     & -0.6580     & -0.5450     \\
                               & OpenCV         & 26.8965      & 27.1316      & 0.8878       & 0.8870       & 62.7130      & 67.9110     & 7.3530       & 6.1690       & 19.0922      & 17.9139     & 0.0000      & 0.0000      \\
                               & StyleGAN 2       & 25.9665      & 26.8775      & 0.8859       & 0.8876       & 59.3050      & 48.5360     & 25.0580      & 15.4620      & 24.3433      & 21.2487     & -5.1980     & -0.0590     \\
                               & Bona Fide      & 29.4893      & 29.5968      & 0.8937       & 0.8952       & 60.2820      & 62.6590     & 43.9160      & 33.9450      & 39.5276      & 34.0680     & -1.3870     & -1.3670     \\ \hline
\multirow{4}{*}{FRGC-Morphs}   & FaceMorpher    & 28.0497      & 27.0605      & 0.8904       & 0.8859       & 71.7660      & 60.9050     & 6.0720       & 6.5410       & 15.0684      & 19.3062     & -2.4470     & -0.4390     \\
                               & OpenCV         & 27.4370      & 27.8318      & 0.8903       & 0.8854       & 69.5100      & 64.9240     & 4.3500       & 3.5600       & 17.1487      & 15.6377     & -3.0620     & -2.0350     \\
                               & StyleGAN 2       & 27.6772      & 29.1389      & 0.8841       & 0.8919       & 62.4120      & 56.3640     & 2.7130       & 10.1530      & 18.7467      & 17.8662     & -2.8610     & -3.6840     \\
                               & Bona Fide      & 28.6661      & 28.8975      & 0.8925       & 0.8879       & 54.6710      & 49.4110     & 5.7000       & 10.9440      & 17.4571      & 19.5861     & -9.6210     & -1.4540     \\ \hline
\multirow{2}{*}{LMA-DRD (D)}   & OpenCV (D)     & 29.0780      & 27.0214      & 0.8901       & 0.8918       & 70.4170      & 61.3760     & 7.0480       & 18.8090      & 15.8341      & 24.1678     & -0.7470     & -5.3790     \\
                               & Bona Fide (D)  & 28.7873      & 29.9078      & 0.8913       & 0.8905       & 64.4650      & 51.9640     & 0.8200       & -2.5890      & 17.5215      & 21.2466     & 0.0000      & -0.1550     \\
\multirow{2}{*}{LMA-DRD (PS)}  & OpenCV (PS)    & 29.1056      & 27.0557      & 0.8906       & 0.8924       & 73.2130      & 67.9370     & 1.2400       & 5.9050       & 14.6985      & 15.8074     & -2.7770     & -0.0440     \\
                               & Bona Fide (PS) & 28.9074      & 29.9588      & 0.8928       & 0.8913       & 69.5030      & 52.0630     & -1.6830      & -6.2090      & 17.5302      & 16.4102     & -1.9700     & -2.0260     \\ \hline
\multirow{3}{*}{MorGAN-Morphs} & OpenCV         & 19.7456      & 14.7414      & 0.8557       & 0.8478       & 44.3260      & 44.6990     & 25.7080      & 18.0130      & 20.9167      & 22.0238     & -5.5650     & -7.8220     \\
                               & MorGAN         & 17.6934      & 18.1678      & 0.8510       & 0.8515       & 47.4040      & 48.4310     & 10.6440      & -0.0210      & 29.4583      & 28.5578     & -5.5000     & -5.5400     \\
                               & Bona Fide      & 16.7996      & 23.6150      & 0.8472       & 0.8630       & 35.7060      & 51.4440     & -1.2140      & 36.9810      & 30.8734      & 26.1947     & -3.0170     & -2.6580     \\ \hline
\end{tabular}
}
}{}
\end{table*}

\begin{table*}
\caption{FDR scores indicate the separability between morphs and bona fide distributions based on the quality metric.\label{tab:FDR} Only intra-dataset comparison is considered here. A larger FDR score indicates a better separability between the morphing attack and bona fide distributions.}
{
\resizebox{\textwidth}{!}{

\begin{tabular}{llllllllllll}
\hline
                              &             & \multicolumn{5}{c}{FIQA}                                          & \multicolumn{5}{c}{IQA}                                      \\ \hline
dataset                      & Morph Type  & MagFace         & SDD-FIQA & SER-FIQ & FaceQnet & rankIQ          & BRISQUE         & DBCNN  & CNNIQA          & UNIQUE & dipIQ  \\ \hline
\multirow{5}{*}{FRLL-Morphs}  & AMSL        & 0.5942          & 0.2021   & 0.0546  & 0.0506   & 0.2063          & \textbf{5.1530} & 4.1005 & 2.7154          & 0.0737 & 0.2901 \\
                              & FaceMorpher & 0.7354          & 0.3761   & 0.0052  & 0.0294   & 0.2157          & 3.5891          & 0.3924 & \textbf{4.3623} & 0.5835 & 0.8133 \\
                              & OpenCV      & 0.8739          & 0.2959   & 0.0079  & 0.0450   & 0.2092          & 1.2141          & 0.0005 & \textbf{4.6174} & 0.0270 & 1.4129 \\
                              & StyleGAN 2    & \textbf{3.6564} & 0.5017   & 0.6746  & 0.0246   & 0.0462          & 0.9034          & 1.0804 & 0.0680          & 0.0744 & 0.3583 \\
                              & WebMorph  & 0.8103          & 0.1516   & 0.0014  & 0.0466   & 0.1146          & \textbf{5.7858} & 1.6706 & 3.3614          & 0.0014 & 0.0532 \\ \hline
\multirow{3}{*}{FERET-Morphs} & FaceMorpher & 0.9094          & 0.0281   & 0.0148  & 0.0092   & 0.2002          & 0.9251          & \textbf{3.3444} & 2.8108 & 0.0098 & 2.1001 \\
                              & OpenCV      & 0.8483          & 0.0197   & 0.0058  & 0.0057   & 0.2071          & 1.9369          & \textbf{5.2332} & 2.6064 & 0.0138 & 1.7226 \\
                              & StyleGAN 2    & \textbf{2.2707} & 0.0000   & 0.2686  & 0.0082   & 0.0760          & 0.5868          & 1.3572 & 0.6977          & 0.0448 & 0.6578 \\ \hline
\multirow{3}{*}{FRGC-Morphs}  & FaceMorpher & 0.4877          & 0.0021   & 0.0064  & 0.0122   & 0.9074          & 0.0009          & 0.0280 & \textbf{0.6465} & 0.0001 & 1.2222 \\
                              & OpenCV      & 0.4492          & 0.0015   & 0.0102  & 0.0109   & \textbf{0.8586} & 0.0428          & 0.1082 & 0.3502          & 0.0009 & 0.5369 \\
                              & StyleGAN 2    & \textbf{2.1508} & 0.0001   & 0.2715  & 0.0278   & 0.2116          & 0.0242          & 0.0855 & 0.0388          & 0.0239 & 0.3793 \\ \hline
LMA-DRD (D)                   & OpenCV (D)  & \textbf{0.7061} & 0.0019   & 0.1309  & 0.0010   & 0.0654          & 0.1646          & 0.0002 & 0.0767          & 0.0017 & 0.0112 \\
LMA-DRD (PS)                  & OpenCV (PS) & \textbf{0.6552} & 0.0006   & 0.1144  & 0.0001   & 0.0423          & 0.0102          & 0.0075 & 0.0648          & 0.0201 & 0.0668 \\ \hline
\multirow{2}{*}{MorGAN}       & OpenCV      & \textbf{0.1562} & 0.0291   & 0.0279  & 0.0375   & 0.0466          & 0.0350          & 0.1349 & 0.0937          & 0.0005 & 0.1157 \\
                              & MorGAN      & \textbf{0.1855} & 0.0268   & 0.1619  & 0.0055   & 0.0194          & 0.4380          & 0.0728 & 0.0050          & 0.0015 & 0.0442 \\ \hline
\end{tabular}
}
}{}
\end{table*}

To draw conclusions on the detectability of the morphing attack by quality measures, we use the presentation attack detection measures defined in ISO/IEC 30107-3 \cite{ISO/IEC_30107-3} as the performance metrics in our work. We used the terms 1) Attack Presentation Classification Error Rate ($APCER$) describing the proportion of attack presentations wrongly classified as bona fides under certain morphing scenarios, 2) the Bona Fide Presentation Classification Error Rate ($BPCER$) describing the proportion of the error of bona fide face images which are wrongly classified as morphed face images, and 3) the Equal Error Rate ($EER$) representing the $APCER$ or $BPCER$ at the operation point where the $APCER$ and $BPCER$ are the same value. We further provide the Average Classification Error Rate $ACER$ describing the average error between $APCER$ and $BPCER$ as given in Equation (\ref{eq:acer}). 

\begin{equation}
    ACER = (APCER+BPCER)/2
    \label{eq:acer}
\end{equation}

The ACER metric is not part of \cite{ISO/IEC_30107-3} standard anymore but helps in giving a single value indicative of the performance. A lower error rate, for all used error metrics, indicates a higher PAD performance.

\subsection{Investigations}

The experiments are designed to address the two main research issues: (1) the effect of different morphing approaches on quality, and (2) the possibility of using these effects to detect morphing attacks.

For the first aspect, the effect of morphing on quality, we (1) investigate the bona fide of each dataset vs. each of the morphing techniques of this dataset. This investigation includes a visual presentation of the quality (different measures) distributions of the bona fide of each dataset alongside the morphing approaches in these datasets. The investigation also includes (2) presenting a quantitative measure of how different are morphing attacks from the bona fide images in each dataset, i.e. the separability FDR measure. 
For the second aspect, the possibility of using the morphing effect on quality to detect attacks, we (1) first look at the possibility to maintain the correct detection of bona fide, even if the bona fide comes from a different source. To do that, we fix the BPCER on each data source and calculate the BPCER on all other data sources. 
Then, (2) we investigate to use this BPCER threshold on each data source to detect the attacks (APCER), despite their differences, in that dataset and all other datasets.
Third (3), after having a detailed look at detecting bona fide and attacks separately (BPCER and APCER), we derive a final overall conclusion by reporting the detailed ACER measures on all datasets and attacks, while fixing the BPCER at each data source individually.

%%%%%%%%%%%%%%%%%%%%%%%
%EER
%%%%%%%%%%%%%%%%%%%%%%%
\begin{table*}
\caption{EER score is presented between the morphs and bona fide distributions using the quality metric as a decision threshold. The quality measures with * indicate that using the inverse of the quality value (lower quality indicates bona fide) leads to a lower average EER and thus is used. Each EER operation point is calculated using the morphing attack and the bona fide scores of the same dataset. In each row of the table, the quality metric with the lowest EER between bona fide and morphing attacks within the same morphing dataset is marked in bold.  Only the \textit{mean} EER value in the last row is averaged across the morphing attacks and morphing datasets alongside the quality measure to determine the best performing 3 measures from both the IQA and FIQA categories, which are in bold.\label{tab:EER}}
{
\resizebox{\textwidth}{!}{
\begin{tabular}{llllllllllll}
\hline
                              &             & \multicolumn{5}{c}{FIQA}                                                  & \multicolumn{5}{c}{IQA}                                               \\ \hline
dataset                      & Morph Type  & MagFace         & SDD-FIQA & SER-FIQ         & FaceQnet & rankIQ*         & BRISQUE         & DBCNN* & CNNIQA*         & UNIQUE & dipIQ*          \\ \hline
\multirow{5}{*}{FRLL-Morphs}  & AMSL        & 0.3122          & 0.3908   & 0.5471          & 0.4524   & 0.3646          & 0.0694          & 0.0883 & \textbf{0.0791} & 0.4437 & 0.3572          \\
                              & FaceMorpher & 0.2856          & 0.3232   & 0.4935          & 0.4755   & 0.3519          & 0.1023          & 0.3077 & \textbf{0.0360} & 0.2807 & 0.2610          \\
                              & OpenCV      & 0.2645          & 0.3620   & 0.4881          & 0.4644   & 0.3407          & 0.2187          & 0.5029 & \textbf{0.0541} & 0.4758 & 0.1384          \\
                              & StyleGAN 2    & \textbf{0.0704} & 0.3437   & 0.2365          & 0.4640   & 0.4272          & 0.2570          & 0.2381 & 0.4313          & 0.5859 & 0.3101          \\
                              & WebMorph  & 0.2826          & 0.3948   & 0.4799          & 0.4480   & 0.3956          & 0.0541          & 0.1867 & \textbf{0.0713} & 0.5061 & 0.4308          \\ \hline
\multirow{3}{*}{FERET-Morphs} & FaceMorpher & 0.2420          & 0.4386   & 0.4575          & 0.4688   & 0.3913          & 0.7391          & 0.8941 & \textbf{0.1080} & 0.5425 & 0.1040          \\
                              & OpenCV      & 0.2514          & 0.4480   & 0.4688          & 0.4669   & 0.3913          & 0.8431          & 0.9452 & \textbf{0.1229} & 0.5501 & 0.1512          \\
                              & StyleGAN 2    & \textbf{0.1399} & 0.4915   & 0.3459          & 0.5198   & 0.4178          & 0.6957          & 0.7977 & 0.2892          & 0.5728 & 0.2647          \\ \hline
\multirow{3}{*}{FRGC-Morphs}  & FaceMorpher & 0.3039          & 0.4761   & 0.5062          & 0.5353   & \textbf{0.2417} & 0.4305          & 0.5280 & 0.3094          & 0.4907 & 0.1815          \\
                              & OpenCV      & 0.3143          & 0.4793   & 0.5156          & 0.5353   & \textbf{0.2448} & 0.5031          & 0.5820 & 0.3703          & 0.4948 & 0.2977          \\
                              & StyleGAN 2    & \textbf{0.1432} & 0.4948   & 0.3268          & 0.5685   & 0.3662          & 0.4886          & 0.4004 & 0.4741          & 0.5207 & 0.3195          \\ \hline
LMA-DRD (D)                   & OpenCV (D)  & \textbf{0.2428} & 0.4783   & 0.3659          & 0.5000   & 0.4239          & 0.3913          & 0.4964 & 0.4529          & 0.4783 & 0.5362          \\
LMA-DRD (PS)                  & OpenCV (PS) & \textbf{0.2509} & 0.4727   & 0.3709          & 0.5091   & 0.4473          & 0.4436          & 0.5127 & 0.4509          & 0.4691 & 0.3891          \\ \hline
\multirow{2}{*}{MorGAN}       & OpenCV      & \textbf{0.3880} & 0.4540   & 0.4470          & 0.5530   & 0.4400          & 0.4520          & 0.3940 & 0.4200          & 0.4970 & 0.6100          \\
                              & MorGAN      & \textbf{0.3750} & 0.4560   & 0.3900          & 0.4780   & 0.5300          & 0.6820          & 0.4150 & 0.4770          & 0.4980 & 0.4540          \\ \hline
\textbf{Mean}                 &             & \textbf{0.2577} & 0.4335   & \textbf{0.4293} & 0.4959   & \textbf{0.3849} & \textbf{0.4247} & 0.4859 & \textbf{0.2778} & 0.4937 & \textbf{0.3203} \\ \hline
\end{tabular}
}
}{}
\end{table*}

%%%%%%%%%%%%%%%%%%%%%%%
%BPCER
%%%%%%%%%%%%%%%%%%%%%%%
\begin{table*}
\caption{BPCER of the morph datasets is given in the table, by fixing the decision threshold so that the BPCER of one morph dataset is at 0.2 (left-most column) and using this threshold to calculate the BPCER for other morphing datasets (second from left column). The bold numbers mark the best performing quality metric within each morphing dataset, which is determined across each row. One can see that despite the change in the bona fide source, thresholding the quality on one dataset does not lead to extremely high BPCER on other datasets in most cases. \label{tab:BPCER}}
\centering
{
\begin{tabular}{llllllll}
\hline
dataset (BF)                 & dataset (BF) & MagFace         & SERFIQ          & rankIQ*         & BRISQUE         & CNNIQA*         & dipIQ*          \\ \hline
\multirow{5}{*}{FRLL-Morphs}  & FERET-Morphs  & 0.3008          & 0.2788          & 0.1415          & 0.9766          & 0.2887          & \textbf{0.0248} \\
                              & FRGC-Morphs   & 0.3098          & 0.3249          & 0.1042          & 0.5513          & 0.6227          & \textbf{0.0736} \\
                              & LMA-DRD (D)   & \textbf{0.1689} & 0.2098          & 0.0027          & 0.4687          & 0.8142          & 0.2207          \\
                              & LMA-DRD (PS)  & 0.2039          & 0.2039          & \textbf{0.0138} & 0.3416          & 0.7438          & 0.2259          \\
                              & MorGAN        & 1.0000          & 1.0000          & 0.0010          & 0.9743          & 0.1316          & \textbf{0.0274} \\ \hline
\multirow{5}{*}{FERET-Morphs} & FRLL-Morphs   & 0.0588          & \textbf{0.1716} & 0.3578          & 0.0000          & 0.0882          & 0.5098          \\
                              & FRGC-Morphs   & 0.1876          & 0.2346          & 0.1734          & \textbf{0.0224} & 0.5463          & 0.2993          \\
                              & LMA-DRD (D)   & 0.0981          & 0.1444          & 0.0245          & \textbf{0.0054} & 0.7486          & 0.5668          \\
                              & LMA-DRD (PS)  & 0.1350          & 0.1460          & 0.0303          & \textbf{0.0028} & 0.6749          & 0.5675          \\
                              & MorGAN        & 1.0000          & 1.0000          & \textbf{0.0023} & 0.1346          & 0.0986          & 0.1764          \\ \hline
\multirow{5}{*}{FRGC-Morphs}  & FRLL-Morphs   & \textbf{0.0686} & 0.1716          & 0.4020          & 0.0000          & 0.0049          & 0.3775          \\
                              & FERET-Morphs  & 0.2088          & 0.1635          & 0.2173          & 0.7481          & \textbf{0.0113} & 0.1203          \\
                              & LMA-DRD (D)   & 0.1035          & 0.1144          & \textbf{0.0245} & 0.1090          & 0.3224          & 0.4142          \\
                              & LMA-DRD (PS)  & 0.1433          & 0.1157          & \textbf{0.0386} & 0.0413          & 0.3691          & 0.4132          \\
                              & MorGAN        & 1.0000          & 1.0000          & \textbf{0.0027} & 0.7348          & 0.0237          & 0.1015          \\ \hline
\multirow{5}{*}{LMA-DRD (D)}  & FRLL-Morphs   & 0.2500          & 0.1961          & 0.7010          & 0.0049          & \textbf{0.0000} & 0.1569          \\
                              & FERET-Morphs  & 0.3418          & 0.2562          & 0.3857          & 0.8662          & \textbf{0.0014} & 0.0191          \\
                              & FRGC-Morphs   & 0.3483          & 0.2952          & 0.4054          & 0.2760          & 0.0821          & \textbf{0.0641} \\
                              & LMA-DRD (PS)  & 0.2314          & 0.1791          & 0.2066          & \textbf{0.0826} & 0.2011          & 0.1983          \\
                              & MorGAN        & 1.0000          & 1.0000          & 0.0307          & 0.8474          & \textbf{0.0107} & 0.0204          \\ \hline
\multirow{5}{*}{LMA-DRD (PS)} & FRLL-Morphs   & 0.1814          & 0.1961          & 0.6961          & 0.0588          & \textbf{0.0000} & 0.1569          \\
                              & FERET-Morphs  & 0.2880          & 0.2725          & 0.3808          & 0.9519          & \textbf{0.0014} & 0.0198          \\
                              & FRGC-Morphs   & 0.2965          & 0.3142          & 0.3997          & 0.4263          & \textbf{0.0815} & 0.0654          \\
                              & LMA-DRD (D)   & \textbf{0.1553} & 0.2071          & 0.1826          & 0.3842          & 0.1913          & 0.2071          \\
                              & MorGAN        & 1.0000          & 1.0000          & 0.0294          & 0.9509          & \textbf{0.0104} & 0.0214          \\ \hline
\multirow{5}{*}{MorGAN}       & FRLL-Morphs   & \textbf{0.0000} & 0.0000          & 0.8971          & 0.0000          & 0.4706          & 0.5490          \\
                              & FERET-Morphs  & \textbf{0.0000} & 0.0000          & 0.6100          & 0.2689          & 0.4416          & 0.2180          \\
                              & FRGC-Morphs   & \textbf{0.0000} & 0.0003          & 0.6681          & 0.0325          & 0.7483          & 0.3331          \\
                              & LMA-DRD (D)   & \textbf{0.0000} & 0.0000          & 0.5967          & 0.0191          & 0.8852          & 0.5913          \\
                              & LMA-DRD (PS)  & \textbf{0.0000} & 0.0028          & 0.5840          & 0.0028          & 0.8347          & 0.6171          \\ \hline
\end{tabular}
}{}
\end{table*}

\begin{figure*}
\centering
\foreach \db in {MagFace,SDD,SERFIQ,FaceQnet,rankIQ}{
\includegraphics[width=0.192\textwidth]{bf_\db.jpg}
}
\foreach \db in {BRISQUE,DBCNN,UNIQUE,CNNIQA,dipIQ}{
\includegraphics[width=0.192\textwidth]{bf_\db.jpg}
}
\includegraphics[width=.75\linewidth]{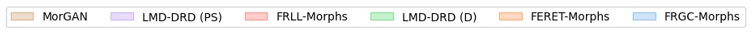}
\caption{The figure depicts the bona fide quality score distributions using kernel density estimation approximation of the 5 morphing datasets according to the proposed quality measures. MorGAN bona fide scores reveal lower quality compared to other bona fide images from other morphing datasets. This figure also provides a slight hint of the quality of the underlying source datasets from which the morphing attacks are created.}
\label{fig:bona_fide_dist}
\end{figure*}

%%%%%%%%%%%%%%%%%%%%%%%
%APCER
%%%%%%%%%%%%%%%%%%%%%%%
\begin{table*}
\caption{The APCER value of the morphing attacks is given in the table at the decision threshold that leads to the BPCER of 0.2 on the dataset in the left-most column. This table aims at showing the morphing detectability using the quality and utility measures within and across morphing attacks and morphing datasets. The bold number shows the lowest APCER error within the same morphing attacks and morphing dataset across different quality measures. Comparison within and across morphing datasets intend to show the MAD performance on known and unknown attacks and its generalizability on unseen morphing images. \label{tab:APCER}}
{
%\begin{adjustbox}{angle=90}
\resizebox{\textwidth}{!}{
\renewcommand{\arraystretch}{1.2}
\begin{tabular}{lllllllllllllllll}
\hline
                              &                & \multicolumn{5}{c}{FRLL-Morphs}                                                         & \multicolumn{3}{c}{FERET-Morphs}                    & \multicolumn{3}{c}{FRGC-Morphs}                     & \multicolumn{1}{c}{LMD-DRD (D)} & \multicolumn{1}{c}{LMD-DRD (PS)} & \multicolumn{2}{c}{MorGAN}        \\ \hline
dataset                      & Quality Metric & AMSL            & FaceMorpher     & OpenCV          & StyleGAN 2        & WebMorph      & FaceMorpher     & OpenCV          & StyleGAN 2        & FaceMorpher     & OpenCV          & StyleGAN 2        & OpenCV (D)                      & OpenCV (PS)                      & OpenCV          & MorGAN          \\ \hline
\multirow{6}{*}{FRLL-Morphs}  & MagFace        & 0.4207          & 0.3560          & 0.3399          & \textbf{0.0123} & 0.3530          & 0.1871          & 0.1871          & \textbf{0.0359} & 0.2925          & 0.3205          & \textbf{0.0373} & 0.3841                          & 0.3127                           & \textbf{0.0000} & \textbf{0.0000} \\
                              & SERFIQ         & 0.9007          & 0.8535          & 0.8559          & 0.2946          & 0.8460          & 0.6919          & 0.7108          & 0.4234          & 0.7376          & 0.7552          & 0.3268          & 0.6449                          & 0.6182                           & \textbf{0.0000} & \textbf{0.0000} \\
                              & rankIQ*        & 0.5857          & 0.5303          & 0.5299          & 0.7995          & 0.6626          & 0.6560          & 0.6635          & 0.7713          & 0.5093          & 0.5104          & 0.8195          & 0.9819                          & 0.9891                           & 0.9990          & 1.0000          \\
                              & BRISQUE        & 0.0239          & 0.0417          & 0.2432          & 0.2938          & 0.0074          & 0.1796          & 0.3743          & 0.1399          & 0.2355          & 0.4118          & 0.3714          & 0.2572                          & 0.5927                           & 0.0060          & 0.1360          \\
                              & CNNIQA*        & \textbf{0.0138} & \textbf{0.0008} & \textbf{0.0049} & 0.7512          & \textbf{0.0008} & \textbf{0.0114} & \textbf{0.0170} & 0.2760          & \textbf{0.0000} & \textbf{0.0041} & 0.1058          & \textbf{0.0580}                 & \textbf{0.1091}                  & 0.7750          & 0.8680          \\
                              & dipIQ*         & 0.9499          & 0.5025          & 0.0958          & 0.4280          & 0.9844          & 0.2684          & 0.3327          & 0.5331          & 0.3506          & 0.6131          & 0.6359          & 0.8007                          & 0.5891                           & 0.9790          & 0.9640          \\ \hline
\multirow{6}{*}{FERET-Morphs} & MagFace        & 0.5756          & 0.5442          & 0.5037          & \textbf{0.0745} & 0.5389          & 0.3214          & 0.3478          & \textbf{0.0813} & 0.4699          & 0.4824          & \textbf{0.0944} & 0.5362                          & 0.4836                           & \textbf{0.0000} & \textbf{0.0000} \\
                              & SERFIQ         & 0.9490          & 0.9157          & 0.9353          & 0.4935          & 0.9132          & 0.7996          & 0.8147          & 0.5633          & 0.8506          & 0.8351          & 0.5083          & 0.7464                          & 0.7527                           & \textbf{0.0000} & \textbf{0.0000} \\
                              & rankIQ*        & 0.3756          & 0.3355          & 0.3235          & 0.5532          & 0.4201          & 0.5803          & 0.5784          & 0.6881          & 0.3371          & 0.3496          & 0.6649          & 0.9457                          & 0.9491                           & 0.9970          & 1.0000          \\
                              & BRISQUE        & 1.0000          & 1.0000          & 1.0000          & 1.0000          & 1.0000          & 0.9924          & 1.0000          & 0.9773          & 1.0000          & 1.0000          & 1.0000          & 0.9855                          & 1.0000                           & 0.8010          & 0.9790          \\
                              & CNNIQA*        & \textbf{0.0925} & \textbf{0.0074} & \textbf{0.0139} & 0.9026          & \textbf{0.0442} & \textbf{0.0227} & \textbf{0.0491} & 0.4291          & \textbf{0.0042} & \textbf{0.0197} & 0.2718          & \textbf{0.1051}                 & \textbf{0.1636}                  & 0.8320          & 0.9090          \\
                              & dipIQ*         & 0.0947          & 0.0106          & 0.0123          & 0.1841          & 0.2940          & 0.0378          & 0.1153          & 0.3138          & 0.0882          & 0.2956          & 0.3506          & 0.5072                          & 0.3055                           & 0.9020          & 0.7820          \\ \hline
\multirow{6}{*}{FRGC-Morphs}  & MagFace        & 0.5549          & 0.5262          & 0.4775          & \textbf{0.0655} & 0.5184          & 0.3100          & 0.3308          & \textbf{0.0699} & 0.4512          & 0.4627          & \textbf{0.0861} & \textbf{0.5145}                 & 0.4655                           & \textbf{0.0000} & \textbf{0.0000} \\
                              & SERFIQ         & 0.9678          & 0.9370          & 0.9533          & 0.6088          & 0.9369          & 0.8488          & 0.8582          & 0.6484          & 0.8859          & 0.8734          & 0.5695          & 0.8116                          & 0.7818                           & \textbf{0.0000} & \textbf{0.0000} \\
                              & rankIQ*        & 0.3232          & 0.2930          & 0.2785          & 0.4943          & \textbf{0.3628} & 0.5595          & 0.5614          & 0.6654          & 0.2956          & \textbf{0.3060} & 0.6307          & 0.9203                          & 0.9345                           & 0.9970          & 1.0000          \\
                              & BRISQUE        & 0.5968          & 0.7692          & 0.9853          & 0.9738          & 0.6830          & 0.7391          & 0.8866          & 0.6333          & 0.9336          & 0.9813          & 0.9741          & 0.7428                          & 0.9636                           & 0.1700          & 0.6340          \\
                              & CNNIQA*        & 0.9489          & 0.8609          & 0.8509          & 0.9992          & 0.9500          & 0.5833          & 0.5614          & 0.8507          & 0.5909          & 0.8154          & 0.9305          & 0.5942                          & 0.5164                           & 0.9660          & 0.9880          \\
                              & dipIQ*         & \textbf{0.2786} & \textbf{0.0458} & \textbf{0.0221} & 0.2651          & 0.6552          & \textbf{0.0756} & \textbf{0.1720} & 0.3951          & \textbf{0.1639} & 0.4108          & 0.4492          & 0.5942                          & \textbf{0.3709}                  & 0.9460          & 0.8620          \\ \hline
\multirow{6}{*}{LMD-DRD (D)}  & MagFace        & 0.3692          & 0.3183          & 0.2842          & \textbf{0.0098} & 0.3129          & \textbf{0.1512} & \textbf{0.1664} & \textbf{0.0265} & 0.2427          & 0.2718          & \textbf{0.0218} & \textbf{0.3225}                 & \textbf{0.2836}                  & \textbf{0.0000} & \textbf{0.0000} \\
                              & SERFIQ         & 0.9172          & 0.8707          & 0.8812          & 0.3576          & 0.8731          & 0.7259          & 0.7316          & 0.4707          & 0.7770          & 0.7790          & 0.3859          & 0.6703                          & 0.6800                           & \textbf{0.0000} & \textbf{0.0000} \\
                              & rankIQ*        & \textbf{0.0887} & \textbf{0.1097} & \textbf{0.1073} & 0.1489          & \textbf{0.1237} & 0.3989          & 0.3913          & 0.4594          & \textbf{0.1110} & \textbf{0.1141} & 0.3226          & 0.6486                          & 0.6727                           & 0.9530          & 0.9900          \\
                              & BRISQUE        & 0.2864          & 0.5025          & 0.9009          & 0.8912          & 0.3292          & 0.5936          & 0.8147          & 0.4934          & 0.8268          & 0.9263          & 0.8932          & 0.6341                          & 0.9236                           & 0.0820          & 0.5010          \\
                              & CNNIQA*        & 0.9995          & 0.9812          & 0.9631          & 1.0000          & 0.9918          & 0.8030          & 0.8034          & 0.9338          & 0.9232          & 0.9678          & 0.9834          & 0.7283                          & 0.7673                           & 0.9850          & 0.9970          \\
                              & dipIQ*         & 0.9651          & 0.5908          & 0.1204          & 0.4509          & 0.9853          & 0.3081          & 0.3516          & 0.5520          & 0.3869          & 0.6463          & 0.6535          & 0.8225                          & 0.6182                           & 0.9800          & 0.9710          \\ \hline
\multirow{6}{*}{LMD-DRD (PS)} & MagFace        & 0.4428          & 0.3781          & 0.3612          & \textbf{0.0188} & 0.3767          & \textbf{0.1947} & \textbf{0.2023} & \textbf{0.0378} & \textbf{0.3133} & 0.3402          & \textbf{0.0415} & 0.4130                          & \textbf{0.3200}                  & \textbf{0.0000} & \textbf{0.0000} \\
                              & SERFIQ         & 0.9053          & 0.8592          & 0.8632          & 0.3142          & 0.8509          & 0.7070          & 0.7164          & 0.4367          & 0.7500          & 0.7604          & 0.3423          & 0.6486                          & 0.6255                           & \textbf{0.0000} & \textbf{0.0000} \\
                              & rankIQ*        & 0.0929          & \textbf{0.1105} & \textbf{0.1089} & 0.1555          & 0.1294          & 0.4026          & 0.3970          & 0.4631          & 0.1131          & \textbf{0.1193} & 0.3330          & 0.6558                          & 0.6873                           & 0.9530          & 0.9900          \\
                              & BRISQUE        & \textbf{0.0782} & 0.1481          & 0.4898          & 0.5589          & \textbf{0.0450} & 0.3043          & 0.5425          & 0.2514          & 0.4346          & 0.6504          & 0.6058          & \textbf{0.3732}                 & 0.7491                           & 0.0170          & 0.2320          \\
                              & CNNIQA*        & 0.9995          & 0.9812          & 0.9631          & 1.0000          & 0.9918          & 0.8030          & 0.8034          & 0.9338          & 0.9263          & 0.9678          & 0.9834          & 0.7319                          & 0.7673                           & 0.9850          & 0.9970          \\
                              & dipIQ*         & 0.9646          & 0.5810          & 0.1171          & 0.4493          & 0.9844          & 0.3043          & 0.3516          & 0.5520          & 0.3828          & 0.6421          & 0.6515          & 0.8225                          & 0.6145                           & 0.9800          & 0.9710          \\ \hline
\multirow{6}{*}{MorGAN}       & MagFace        & 1.0000          & 1.0000          & 1.0000          & 1.0000          & 1.0000          & 1.0000          & 1.0000          & 1.0000          & 1.0000          & 1.0000          & 1.0000          & 1.0000                          & 1.0000                           & \textbf{0.6100} & \textbf{0.6620} \\
                              & SERFIQ         & 1.0000          & 1.0000          & 1.0000          & 1.0000          & 1.0000          & 1.0000          & 1.0000          & 1.0000          & 1.0000          & 1.0000          & 1.0000          & 1.0000                          & 1.0000                           & 0.7700          & 0.6800          \\
                              & rankIQ*        & 0.0211          & 0.0417          & 0.0467          & \textbf{0.0336} & \textbf{0.0590} & 0.2136          & 0.2079          & 0.2344          & 0.0207          & 0.0322          & 0.0871          & 0.2681                          & 0.3055                           & 0.7000          & 0.8660          \\
                              & BRISQUE        & 1.0000          & 1.0000          & 1.0000          & 1.0000          & 1.0000          & 0.9924          & 0.9981          & 0.9622          & 1.0000          & 1.0000          & 1.0000          & 0.9819                          & 1.0000                           & 0.7290          & 0.9630          \\
                              & CNNIQA*        & \textbf{0.0000} & \textbf{0.0000} & \textbf{0.0008} & 0.3674          & 0.0000          & \textbf{0.0000} & \textbf{0.0095} & \textbf{0.1323} & \textbf{0.0000} & \textbf{0.0000} & \textbf{0.0197} & \textbf{0.0362}                 & \textbf{0.0764}                  & 0.6890          & 0.7850          \\
                              & dipIQ*         & 0.0699          & 0.0065          & 0.0082          & 0.1702          & 0.2506          & 0.0340          & 0.1040          & 0.2987          & 0.0768          & 0.2604          & 0.3133          & 0.4601                          & 0.2655                           & 0.8850          & 0.7340          \\ \hline
\end{tabular}
}
%\end{adjustbox}
}{}
\end{table*}

%%%%%%%%%%%%%%%%%%%%%%%
%ACER
%%%%%%%%%%%%%%%%%%%%%%%
\begin{table*}
\caption{The ACER of each morphing method and associated bona fide is listed in the table at the decision threshold that leads to the BPCER of 0.2 on the dataset in the left-most column. The ACER shows a better representation of the trade-off between false positives and false negatives. The bold figures indicate the best performing quality metric within each attack type and morphing dataset. The last column shows the average ACER across all morphing attacks and thus shows the average performance for one quality metric as an MAD. In the last column, all quality measures with an ACER below 0.3 are stressed in bold to emphasize their overall performance across different MADs. \label{tab:ACER}}
{
\resizebox{\textwidth}{!}{
%\begin{adjustbox}{angle=90}
\renewcommand{\arraystretch}{1.3}
\begin{tabular}{lllllllllllllllll|l}
\hline
                              &                & \multicolumn{5}{c}{FRLL-Morphs}                                                         & \multicolumn{3}{c}{FERET-Morphs}                    & \multicolumn{3}{c}{FRGC-Morphs}                     & \multicolumn{1}{c}{LMD-DRD (D)} & \multicolumn{1}{c}{LMD-DRD (PS)} & \multicolumn{2}{c|}{MorGAN}       & Mean            \\ \hline
dataset                      & Quality Metric & AMSL            & FaceMorpher     & OpenCV          & StyleGAN 2        & WebMorph      & FaceMorpher     & OpenCV          & StyleGAN 2        & FaceMorpher     & OpenCV          & StyleGAN 2        & OpenCV (D)                      & OpenCV (PS)                      & OpenCV          & MorGAN          &                 \\ \hline
\multirow{6}{*}{FRLL-Morphs}  & MagFace        & 0.3108          & 0.2785          & 0.2704          & \textbf{0.1066} & 0.2770          & 0.2440          & 0.2440          & \textbf{0.1683} & 0.3011          & 0.3151          & \textbf{0.1736} & \textbf{0.2765}                 & \textbf{0.2583}                  & 0.5000          & 0.5000          & \textbf{0.2816} \\
                              & SERFIQ         & 0.5508          & 0.5272          & 0.5284          & 0.2478          & 0.5235          & 0.4854          & 0.4948          & 0.3511          & 0.5312          & 0.5400          & 0.3258          & 0.4274                          & 0.4110                           & 0.5000          & 0.5000          & 0.4629          \\
                              & rankIQ*        & 0.3934          & 0.3656          & 0.3654          & 0.5002          & 0.4318          & 0.3987          & 0.4025          & 0.4564          & 0.3068          & 0.3073          & 0.4619          & 0.4923                          & 0.5014                           & 0.5000          & 0.5005          & 0.4256          \\
                              & BRISQUE        & 0.1124          & 0.1214          & 0.2221          & 0.2474          & 0.1042          & 0.5781          & 0.6755          & 0.5583          & 0.3934          & 0.4816          & 0.4613          & 0.3630                          & 0.4672                           & 0.4901          & 0.5551          & 0.3887          \\
                              & CNNIQA*        & \textbf{0.1074} & \textbf{0.1009} & \textbf{0.1029} & 0.4761          & \textbf{0.1009} & 0.1501          & \textbf{0.1529} & 0.2824          & 0.3113          & \textbf{0.3134} & 0.3642          & 0.4361                          & 0.4264                           & \textbf{0.4533} & 0.4998          & \textbf{0.2852} \\
                              & dipIQ*         & 0.5754          & 0.3517          & 0.1484          & 0.3145          & 0.5927          & \textbf{0.1466} & 0.1787          & 0.2789          & \textbf{0.2121} & 0.3433          & 0.3547          & 0.5107                          & 0.4075                           & 0.5032          & \textbf{0.4957} & 0.3609          \\ \hline
\multirow{6}{*}{FERET-Morphs} & MagFace        & 0.3172          & 0.3015          & 0.2813          & \textbf{0.0666} & 0.2989          & 0.2608          & 0.2741          & \textbf{0.1408} & 0.3287          & 0.3350          & \textbf{0.1410} & \textbf{0.3172}                 & \textbf{0.3093}                  & 0.5000          & 0.5000          & \textbf{0.2914} \\
                              & SERFIQ         & 0.5603          & 0.5436          & 0.5534          & 0.3325          & 0.5424          & 0.4992          & 0.5068          & 0.3811          & 0.5426          & 0.5348          & 0.3715          & 0.4454                          & 0.4494                           & 0.5000          & 0.5000          & 0.4842          \\
                              & rankIQ*        & 0.3667          & 0.3467          & 0.3407          & 0.4555          & 0.3890          & 0.3910          & 0.3901          & 0.4449          & 0.2552          & \textbf{0.2615} & 0.4191          & 0.4851                          & 0.4897                           & 0.4997          & 0.5012          & 0.4024          \\
                              & BRISQUE        & 0.5000          & 0.5000          & 0.5000          & 0.5000          & 0.5000          & 0.5946          & 0.5984          & 0.5870          & 0.5112          & 0.5112          & 0.5112          & 0.4955                          & 0.5014                           & 0.4678          & 0.5568          & 0.5223          \\
                              & CNNIQA*        & \textbf{0.0903} & \textbf{0.0478} & \textbf{0.0511} & 0.4954          & \textbf{0.0662} & \textbf{0.1087} & \textbf{0.1219} & 0.3119          & 0.2752          & 0.2830          & 0.4090          & 0.4269                          & 0.4193                           & \textbf{0.4653} & 0.5038          & \textbf{0.2717} \\
                              & dipIQ*         & 0.3023          & 0.2602          & 0.2610          & 0.3470          & 0.4019          & 0.1187          & 0.1574          & 0.2567          & \textbf{0.1938} & 0.2975          & 0.3250          & 0.5370                          & 0.4365                           & 0.5392          & \textbf{0.4792} & 0.3275          \\ \hline
\multirow{6}{*}{FRGC-Morphs}  & MagFace        & 0.3118          & 0.2974          & 0.2731          & \textbf{0.0670} & \textbf{0.2935} & 0.2594          & 0.2698          & \textbf{0.1394} & 0.3256          & 0.3313          & \textbf{0.1430} & \textbf{0.3090}                 & \textbf{0.3044}                  & 0.5000          & 0.5000          & \textbf{0.2883} \\
                              & SERFIQ         & 0.5697          & 0.5543          & 0.5624          & 0.3902          & 0.5543          & 0.5061          & 0.5109          & 0.4059          & 0.5429          & 0.5367          & 0.3847          & 0.4630                          & 0.4488                           & 0.5000          & 0.5000          & 0.4953          \\
                              & rankIQ*        & 0.3626          & 0.3475          & 0.3402          & 0.4481          & 0.3824          & 0.3884          & 0.3894          & 0.4413          & 0.2479          & \textbf{0.2531} & 0.4154          & 0.4724                          & 0.4866                           & 0.4998          & 0.5013          & 0.3984          \\
                              & BRISQUE        & \textbf{0.2984} & 0.3846          & 0.4926          & 0.4869          & 0.3415          & 0.7436          & 0.8173          & 0.6907          & 0.5671          & 0.5909          & 0.5873          & 0.4259                          & 0.5025                           & \textbf{0.4524} & 0.6844          & 0.5377          \\
                              & CNNIQA*        & 0.4769          & 0.4329          & 0.4279          & 0.5020          & 0.4775          & 0.2973          & 0.2864          & 0.4310          & 0.3955          & 0.5078          & 0.5653          & 0.4583                          & 0.4428                           & 0.4949          & 0.5059          & 0.4468          \\
                              & dipIQ*         & 0.3280          & \textbf{0.2116} & \textbf{0.1998} & 0.3213          & 0.5163          & \textbf{0.0980} & \textbf{0.1462} & 0.2577          & \textbf{0.1817} & 0.3052          & 0.3244          & 0.5042                          & 0.3921                           & 0.5238          & \textbf{0.4818} & 0.3194          \\ \hline
\multirow{6}{*}{LMD-DRD (D)}  & MagFace        & 0.3096          & 0.2842          & 0.2671          & \textbf{0.1299} & 0.2814          & 0.2465          & 0.2541          & \textbf{0.1841} & 0.2955          & 0.3100          & \textbf{0.1850} & \textbf{0.2620}                 & \textbf{0.2575}                  & 0.5000          & 0.5000          & \textbf{0.2844} \\
                              & SERFIQ         & 0.5567          & 0.5334          & 0.5387          & 0.2768          & 0.5346          & 0.4910          & 0.4939          & 0.3634          & 0.5361          & 0.5371          & 0.3406          & 0.4332                          & 0.4295                           & 0.5000          & 0.5000          & 0.4710          \\
                              & rankIQ*        & 0.3949          & 0.4053          & 0.4041          & 0.4250          & 0.4123          & 0.3923          & 0.3885          & 0.4225          & 0.2582          & \textbf{0.2598} & 0.3640          & 0.4237                          & 0.4397                           & 0.4919          & 0.5104          & 0.3995          \\
                              & BRISQUE        & \textbf{0.1457} & \textbf{0.2537} & 0.4529          & 0.4480          & \textbf{0.1671} & 0.7299          & 0.8405          & 0.6798          & 0.5514          & 0.6012          & 0.5846          & 0.4178                          & 0.5031                           & \textbf{0.4647} & 0.6742          & 0.5009          \\
                              & CNNIQA*        & 0.4998          & 0.4906          & 0.4816          & 0.5000          & 0.4959          & 0.4022          & 0.4024          & 0.4676          & 0.5026          & 0.5250          & 0.5327          & 0.4639                          & 0.4842                           & 0.4978          & 0.5038          & 0.4833          \\
                              & dipIQ*         & 0.5610          & 0.3738          & \textbf{0.1386} & 0.3039          & 0.5711          & \textbf{0.1636} & \textbf{0.1854} & 0.2855          & \textbf{0.2255} & 0.3552          & 0.3588          & 0.5107                          & 0.4083                           & 0.5002          & \textbf{0.4957} & 0.3624          \\ \hline
\multirow{6}{*}{LMD-DRD (PS)} & MagFace        & 0.3121          & 0.2797          & 0.2713          & \textbf{0.1001} & 0.2791          & 0.2414          & 0.2452          & \textbf{0.1629} & 0.3049          & 0.3184          & \textbf{0.1690} & \textbf{0.2842}                 & \textbf{0.2606}                  & 0.5000          & 0.5000          & \textbf{0.2819} \\
                              & SERFIQ         & 0.5507          & 0.5277          & 0.5297          & 0.2552          & 0.5235          & 0.4897          & 0.4945          & 0.3546          & 0.5321          & 0.5373          & 0.3283          & 0.4278                          & 0.4133                           & 0.5000          & 0.5000          & 0.4642          \\
                              & rankIQ*        & 0.3945          & 0.4033          & 0.4025          & 0.4258          & 0.4127          & 0.3917          & 0.3889          & 0.4219          & 0.2564          & \textbf{0.2595} & 0.3664          & 0.4192                          & 0.4428                           & 0.4912          & 0.5097          & 0.3991          \\
                              & BRISQUE        & \textbf{0.0685} & \textbf{0.1035} & 0.2743          & 0.3089          & \textbf{0.0519} & 0.6281          & 0.7472          & 0.6016          & 0.4305          & 0.5383          & 0.5160          & 0.3787                          & 0.4737                           & \textbf{0.4840} & 0.5915          & 0.4131          \\
                              & CNNIQA*        & 0.4998          & 0.4906          & 0.4816          & 0.5000          & 0.4959          & 0.4022          & 0.4024          & 0.4676          & 0.5039          & 0.5247          & 0.5324          & 0.4616                          & 0.4828                           & 0.4977          & 0.5037          & 0.4831          \\
                              & dipIQ*         & 0.5607          & 0.3689          & \textbf{0.1370} & 0.3031          & 0.5707          & \textbf{0.1621} & \textbf{0.1857} & 0.2859          & \textbf{0.2241} & 0.3537          & 0.3584          & 0.5148                          & 0.4064                           & 0.5007          & \textbf{0.4962} & 0.3618          \\ \hline
\multirow{6}{*}{MorGAN}       & MagFace        & 0.5000          & 0.5000          & 0.5000          & 0.5000          & 0.5000          & 0.5000          & 0.5000          & 0.5000          & 0.5000          & 0.5000          & 0.5000          & 0.5000                          & 0.5000                           & \textbf{0.4050} & \textbf{0.4310} & 0.4890          \\
                              & SERFIQ         & 0.5000          & 0.5000          & 0.5000          & 0.5000          & 0.5000          & 0.5000          & 0.5000          & 0.5000          & 0.5002          & 0.5002          & 0.5002          & 0.5000                          & 0.5014                           & 0.4847          & 0.4397          & 0.4950          \\
                              & rankIQ*        & 0.4591          & 0.4694          & 0.4719          & 0.4653          & 0.4780          & 0.4118          & 0.4090          & 0.4222          & 0.3444          & 0.3501          & 0.3776          & \textbf{0.4324}                 & 0.4447                           & 0.4500          & 0.5330          & 0.4345          \\
                              & BRISQUE        & 0.5000          & 0.5000          & 0.5000          & 0.5000          & 0.5000          & 0.6307          & 0.6335          & 0.6156          & 0.5163          & 0.5163          & 0.5163          & 0.5005                          & 0.5014                           & 0.4645          & 0.5815          & 0.5317          \\
                              & CNNIQA*        & \textbf{0.2353} & \textbf{0.2353} & \textbf{0.2357} & 0.4190          & \textbf{0.2353} & 0.2208          & 0.2255          & 0.2870          & 0.3742          & 0.3742          & 0.3840          & 0.4607                          & 0.4555                           & 0.4456          & 0.4936          & 0.3387          \\
                              & dipIQ*         & 0.3095          & 0.2778          & 0.2786          & \textbf{0.3596} & 0.3998          & \textbf{0.1260} & \textbf{0.1610} & \textbf{0.2583} & \textbf{0.2049} & \textbf{0.2967} & \textbf{0.3232} & 0.5257                          & \textbf{0.4413}                  & 0.5427          & 0.4672          & \textbf{0.3314} \\ \hline
\end{tabular}
}
%\end{adjustbox}
}{}
\end{table*}

\section{Results and discussion} \label{ch:discussion}
The two main research gaps that this work addresses are (1) the effect of face morphing on image quality and utility, and (2) the detectability of these different morphing attacks based on image quality and utility. 
The experimental outcome regarding these two issues is discussed based on the detailed results in this section.

\subsection{The effect of morphing on quality}

\paragraph{How different are the quality scores distributions of bona fide and morphed samples?}

To answer this question, we look at the quality score distributions produced by each of the 10 used quality measures. 
The distribution comparison is within each dataset bona fide and morphing attacks.
Figure \ref{fig:distributions} shows the quality score distributions of different morphing approaches versus its bona fide distribution within each of the 5 morphing datasets. Each row represents one IQ or FIQ measure. 
Observing the quality of morphing attacks and bona fide as score distributions in Figure \ref{fig:distributions}, we see that different morphing attacks shift the value of IQ/FIQ compared to bona fide images for most quality measures. 

In the FRLL-Morphs, landmark-based attacks showed closer quality distribution to the respective bona fide images, while StyleGAN 2 generated morphs clearly showed lower quality compared to landmark-based morphs measured by quality measures such as MagFace, SDD-FIQA, SER-FIQ, and BRISQUE. The IQA measures like UNIQUE, CNNIQA, and dipIQ even signed slightly higher quality for StyleGAN 2 generated morphing attacks than bona fide distributions. FaceQnet seems to have more difficulty distinguishing the quality difference between the morphs and bona fide images.
Similar conclusions are to be drawn for both FERET-Morphs and FRGC-Morphs, where StyleGAN 2 created morphs show lower image quality measured by MagFace, SERFIQ, BRISQUE, and DBCNN. In FERET-Morphs, DBCNN assigns the bona fide images better quality compared to other morphing attacks of different kinds. CNNIQA on the contrary assigns lower quality values to bona fide images compared to morphing attacks. 

For digital and re-digitized attacks in the LMA-DRD dataset, only MagFace, SER-FIQ, rankIQ, and BRISQUE clearly show a quality degradation of morphing attacks towards bona fide images. CNNIQA even seems to have slightly higher quality for the digital and re-digitized morphing attacks compared to bona fide. Looking at the MorGAN dataset, consistent findings as in \cite{DBLP:conf/iciap/DebiasiDSRSBKU19, DBLP:conf/btas/DamerS0K18} are observed, where we could confirm that IQA measures, such as CNNIQA and dipIQ have similar image quality between MorGAN generated attacks and bona fide, while BRISQUE shows the opposite conclusion. SER-FIQ and BRISQUE put the quality of the landmark-based attacks closer to the bona fide images. 
However, as notice in Figure \ref{fig:distributions}, the quality of MorGAN attacks and bona fide, for almost all FIQA and IQA, is lower than other datasets (keep in mind that some quality measures has reversed interpretation as presented in Table \ref{tab:quality_metrics}). As bona fide samples of the MorGAN data are also of relative low quality, this must be strongly influenced by the low resolution of MorGAN data (64x64 pixels). This strong influence of resolution is evident as it is the only major factor that differentiate between he bona fide samples of all other datasets and the ones of the MorGAN dataset.

\paragraph{Are the quality distributions of bona fide and attack samples separable? and how separable are they?}
Table \ref{tab:FDR} shows the FDR value between each morphing attack to the bona fide quality distribution within each morphing dataset. A large FDR value indicates a high separability between both distributions and thus indicates the possibility of using this quality metric for detecting this kind of morphing attack among the bona fide images.

One finding based on the FDR in Table \ref{tab:FDR} is that we observe MagFace having relatively high separability for GAN-based morphing attacks (such as StyleGAN 2, MorGAN, and even for re-digitized morphing attacks in the LMA-DRD dataset), while rankIQ, BRISQUE, and CNNIQA outperform mostly in separating landmark-based morphing attacks. The highest FDR values in Table \ref{tab:FDR} are marked in bold for improved visibility. This finding might be justified by the nature of the IQA measures looking at the general image quality but having trouble focusing on learning the specifical GAN-based artifacts in a face image. BRISQUE and CNNIQA as representative quality measures of the IQA family focus more on the visible artifacts in the morphed images and less on realistic artifacts introduced by GAN-based approaches. These visible artifacts are easier to be introduced by landmark-based morphing techniques using alpha blending of two bona fide sources like in the case of FaceMorpher, OpenCV, and WebMorph in the FRLL-Morphs, FERET-Morphs, and FRGC-Morphs datasets as visualized in Figure \ref{fig:samples_morph}. Another CNN-based quality measure using multitask learning (DBCNN) is even outperforming CNNIQA in FERET-Morphs showing stronger separability for landmark-based morphs to bona fide and as well as for StyleGAN 2 generated morphs (but still inferior to MagFace in this case).

\paragraph{Are morphing attacks with relatively low/high IQ also lead to low/high FIQ?}

Several works in \cite{DBLP:conf/wacv/FuCHD22,ma2017end} pointed out that the FIQ measures such as MagFace \cite{meng2021magface}, SDD-FIQA \cite{DBLP:journals/corr/abs-2103-05977}, SER-FIQ \cite{terhorst2020ser}, and CR-FIQA \cite{DBLP:journals/corr/abs-2112-06592} are highly correlated with the face utility as defined in ISO/IEC 29794-1 \cite{ISOIEC29794-1}. A recent work \cite{DBLP:conf/wacv/FuCHD22} has shown that for normal FR samples, IQ measures also correlate to utility but to a much lower degree than FIQ. That work also showed low correlation between the decisions of IQA and FIQA.
Knowing that IQA intends to measure the perceptual quality and FIQA intends to measure the image utility, one can see from the quality distributions in Figure \ref{fig:distributions} that not only the IQ is effected by morphing, but also the utility related FIQ measures, more clearly for MagFace than others.

\begin{figure*}
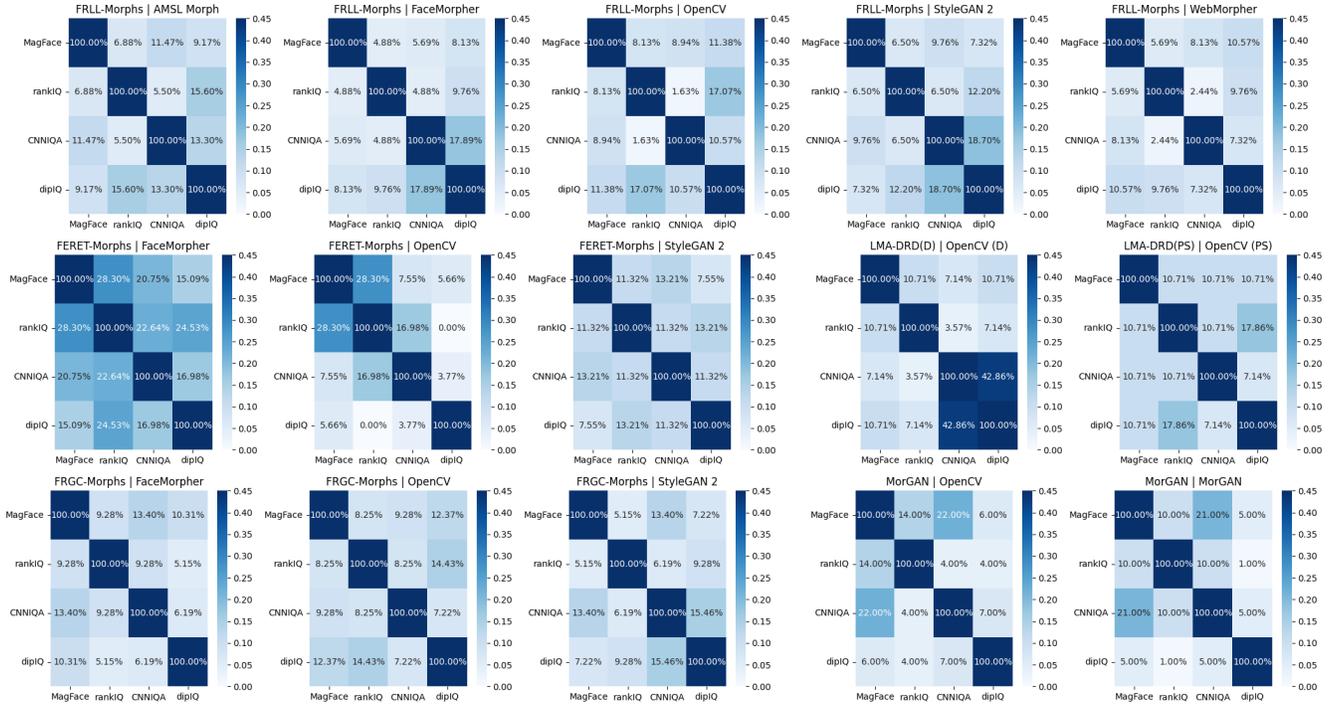

\centering
\foreach \method in {high}{
\foreach \db in {frll}{
\foreach \morph in {amsl,facemorpher,opencv,stylegan,webmorph}
{\includegraphics[width=0.19\textwidth]{conf_matrix_\db_\morph_\method.png}}
}
\foreach \db in {feret}{
\foreach \morph in {facemorpher,opencv,stylegan}
{\includegraphics[width=0.19\textwidth]{conf_matrix_\db_\morph_\method.png}}
}
\foreach \db in {LMA-DRD}{
\foreach \morph in {digital,ps}
{\includegraphics[width=0.19\textwidth]{conf_matrix_\db_\morph_\method.png}}
}
\foreach \db in {frgc}{
\foreach \morph in {facemorpher,opencv,stylegan}
{\includegraphics[width=0.19\textwidth]{conf_matrix_\db_\morph_\method.png}}
}
\foreach \db in {MorGAN}{
\foreach \morph in {LMA, MorGAN}
{\includegraphics[width=0.19\textwidth]{conf_matrix_\db_\morph_\method.png}}
}}
\caption{The overlap ratio between the morphing attack samples designated to be among the highest 10\% qualities (in their respective datasets) as ranked by different pairs of FIQA and IQA methods (on the X and Y axes). MagFace and RankIQ are FIQA methods, and CNNIQA and deepIQ are IQA methods. This indicates the morphing attacks with relatively high IQ does not necessary lead to high FIQ.}
\label{fig:conf_high}
\end{figure*}

\begin{figure*}
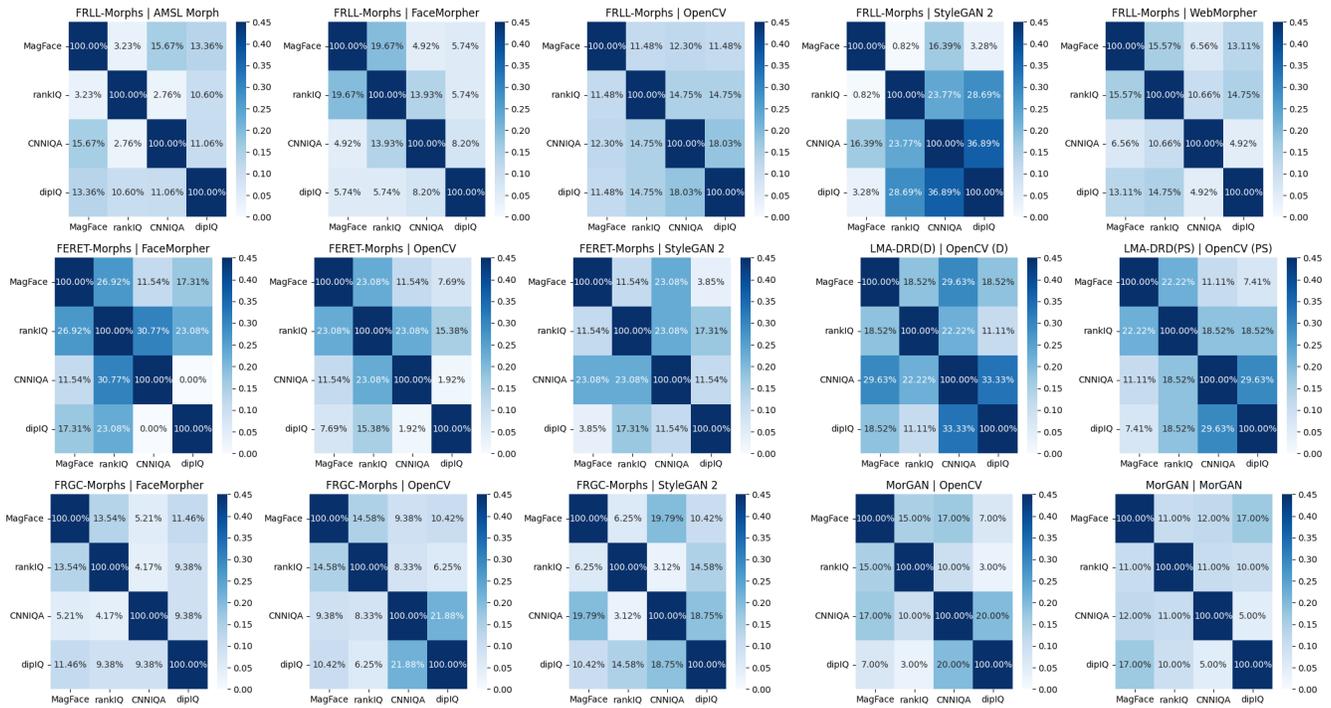

\centering
\foreach \method in {low}{
\foreach \db in {frll}{
\foreach \morph in {amsl,facemorpher,opencv,stylegan,webmorph}
{\includegraphics[width=0.19\textwidth]{conf_matrix_\db_\morph_\method.png}}
}
\foreach \db in {feret}{
\foreach \morph in {facemorpher,opencv,stylegan}
{\includegraphics[width=0.19\textwidth]{conf_matrix_\db_\morph_\method.png}}
}
\foreach \db in {LMA-DRD}{
\foreach \morph in {digital,ps}
{\includegraphics[width=0.19\textwidth]{conf_matrix_\db_\morph_\method.png}}
}
\foreach \db in {frgc}{
\foreach \morph in {facemorpher,opencv,stylegan}
{\includegraphics[width=0.19\textwidth]{conf_matrix_\db_\morph_\method.png}}
}
\foreach \db in {MorGAN}{
\foreach \morph in {LMA, MorGAN}
{\includegraphics[width=0.19\textwidth]{conf_matrix_\db_\morph_\method.png}}
}}
\caption{The overlap ratio between the morphing attack samples designated to be among the lowest 10\% qualities (in their respective datasets) as ranked by different pairs of FIQA and IQA methods (on the X and Y axes). MagFace and RankIQ are FIQA methods, and CNNIQA and deepIQ are IQA methods. This indicates the morphing attacks with relatively low IQ does not necessary lead to low FIQ.}
\label{fig:conf_low}
\end{figure*}

To quantitatively represent the correlation between IQA and FIQA, we look into the lowest and highest quality samples for a set of FIQA and IQA measures and measure the ratio of the shared (overlapped) samples in each of these groups (high or low quality).
To calculate this correlation between quality estimation methods, we calculate the attack samples overlap ratio between the samples of the lowest quality (10\% of the data) between every pair of quality estimation methods, and the same for the 10\% of the highest quality. A large overlapping ratio indicates a larger reasoning similarity between the considered pair of methods.
These ratios are presented in Figure \ref{fig:conf_high} for the top 10\% qualities and in Figure \ref{fig:conf_low} for the bottom 10\% quality, for the 2 FIQA and 2 IQA methods that showed the largest separability between morphing attacks and bona fide samples, i.e. MagFace and RankIQ from FIQA methods, and CNNIQA and dipIQ from IQA methods. 
For most morphing attacks, the correlation is slightly larger within FIQA methods (i.e. between MagFace and RankIQ) and within IQA methods (i.e. CNNIQA and dipIQ) than it is between FIQA and IQA methods. This is not always true, as CNNIQA and MagFace show a relatively higher correlation when assessing StyleGAN2 morphs. However, the correlation between IQ and FIQ measures is relatively low, indicating that, as proven previously on bona fide samples in \cite{DBLP:conf/wacv/FuCHD22}, an image with low perceptual quality does not necessarily lead to a low FR utility and vise versa. A clear example of that is applying FR on masked faces \cite{DBLP:journals/iet-bmt/DamerBSKK22,DBLP:journals/pr/FangDKK22}, where an image of a masked face can be of high perceptual quality, but it is of low utility given the mask occlusion \cite{DBLP:conf/fgr/FuKD21}.

So far, our investigations showed that some quality measures are clearly effected by the morphing process. We have seen also that the different morphing techniques effect these quality measures differently. A major outcome is that some quality measures, such as MagFace, do result in a high separability between bona fide samples and the different attacks. This will be leveraged in the next section to investigate how can this separability be used to detect morphing attacks and how generalizable is this detection over multiple morphing attack methods and bona fide sources. A side observation is that morphing attacks with relatively low/high IQ does not necessary lead to low/high FIQ, confirming previous works conclusions on normal FR tasks \cite{DBLP:conf/wacv/FuCHD22}.

\subsection{The unsupervised detectability of morphing attacks by quality measures}

%EER
%but bona fide is also different

%so we need to look at BPCER and APCER -> ACER

In this section, we start by looking at the detectability within each dataset, motivated by the bona fide-attack separability presented in the last section. 
Then, motivated by the differences in the quality distributions of the different bona fide sources, we investigate the generalizability of detecting morphing attacks/bona fide by quality over different bona fide sources and attack methods. 

\paragraph{If quality measures were used as an MAD detection score, what would be the detection error within each of the investigated datasets?}
Table \ref{tab:EER} shows the EER for each morphing technique in each dataset by using the quality measures as MAD scores. It is to note the $\star$ behind the quality and utility measure in the table. This $\star$ indicates the reversed ordering of the quality measure to its original meaning as listed in Table \ref{tab:quality_metrics}. Looking at the quality distribution for dipIQ, we observed that the bona fide quality is lower compared to the morphing attacks. Therefore, in order to use this quality metric as an MAD score, we have to reverse the quality scores to make the score thresholds consistent with other quality measures considered in this work. The same is for rankIQ, DBCNN, and CNNIQA measures. The smaller the EER, the better is this quality metric suited for MAD task. The numbers in bold indicate the lowest EER for each attack type across all used quality measures within the same morphing dataset. The last row in the table shows the \textit{mean} overall EER scores for each quality metric across all morphing types and datasets. This overall performance provided us the indication, which quality metric performs generally well on most of the MAD tasks. The best 3 performing quality measures from the FIQA and IQA categories are selected for further detailed cross-datasets investigations. Therefore, for the rest of the experiments, 6 FIQA/IQA methods will be considered, namely the MagFace, SER-FIQ, rankIQ, BRISQUE, CNNIQA, and dipIQ.

Based on the EER values in Table \ref{tab:EER}, and consistent with the previously discussed FDR values,  MagFace shows the lowest EER for StyleGAN 2 generated morphs in the FRLL-Morphs, while inverted CNNIQA shows the lowest EER for all landmark-based morphs within this dataset. The corresponding finding is to be observed for FERET-Morphs. Regarding the FRGC-Morphs, the inverted rankIQ surprisingly shows lower EER for FaceMorpher and OpenCV morphs compared to the inverted CNNIQA. As for the LMA-DRD datasets and the MorGAN-dataset, MagFace constantly outperform the other quality measures in terms of EER by a large margin to further consolidate its superior ability to be used as a decision metric in the detectability of morphing attacks. We observed that MagFace and the inverted CNNIQA show overall low EER across all morphing attacks and datasets.

\paragraph{How different are the qualities of different bona fide samples? and how does that effect the correct detectability of bona fide?}

Figure \ref{fig:bona_fide_dist} depicts the quality distributions of all the bona fide images across the 5 data sources using all 10 selected quality measures. Since the MorGAN dataset has been originated from lower image resolution, the quality of this dataset is visibly lower compared to other bona fide sources. This is shown by all FIQA measures. MagFace, SDD-FIQA, and SER-FIQ clearly placed the quality of the MorGAN bona fide images considerably lower than bona fide images in other data sources. Only CNNIQA, DBCNN, and UNIQUE did not place the  MorGAN bona fide images as far from other bona fide images. 
Besides bona fide images in MorGAN dataset, other datasets showed slightly different quality distributions, however, not consistently over all quality measures. FRLL bona fide images consistently showed tight quality distributions in comparison to other data sources, which might be due to the highly controlled capture environment in FRLL.
The FERET bona fide samples showed relatively shifted quality distributions in comparison to other data sources, especially with the BRISQUE, DBCNN, and CNNIQA measures, which might be due to the fact that the FERET dataset was captured in the early 1990s with relatively outdated cameras, and thus differences in the resulting images.
In general, putting MorGAN (low resolution and thus low utility) aside, all bona fide sources showed similar FIQA distributions despite the differences in the perceived quality.

Given that the quality values can be used as an MAD score, setting the BPCER value on one bona fide source, we investigate what would be the resulting BPCER on the other data sources. To show that, Table \ref{tab:BPCER} listed the BPCER values for all other bona fide images of the other data sources (second column) when fixing the decision threshold so the BPCER of the data source (in the first column) is at 0.2. Along each row, the lowest BPCER (with the most effective quality measure as an MAD score) is marked in bold. Under most settings, MagFace, the inverted rankIQ, BRISQUE, inverted CNNIQA, and dipIQ show low BPCER on other bona fide sources when counting the occurrences of minimum BPCER in bold across settings. We can see from Table \ref{tab:BPCER}  that despite the change in the bona fide source, thresholding the quality on one source does not lead to extremely high BPCER on other data sources in most cases. If we neglected the unrealistic MorGAN bona fide, the FIQA methods perform significantly better in detecting bona fide than IQA methods.

\paragraph{How detectable are the different morphing attacks by quality when fixing the decision threshold on diverse bona fide sources (i.e. BPCER)?}

To answer this, we fix the BPCER, each time on the bona fide of one of the five data sources, and calculate the APCER of each of the attack types in each of the morphing datasets.
Table \ref{tab:APCER} shows the APCER of the morphing attacks when fixing the BPCER of the individual bona fide scores at 0.2 for the bona fide source in the first column. Considering the first row, the experimental setting is to fix e.g. the decision threshold using a quality measure for the bona fide distribution of the FRLL-Morphs at BPCER=0.2 and use this decision threshold to determine the APCER value of AMSL, FaceMorpher, OpenCV, StyleGAN 2, and WebMorph attacks in the FRLL-Morphs. The APCER value is not only determined for all morphing attack types within the same morphing dataset but also across all other attacks in the other morphing datasets as well. This is targeted at studying the effectiveness of using IQ and FIQ measures for MAD on unknown morphing attacks and bona fide sources. 

The bold number in Table \ref{tab:APCER} shows the lowest APCER within the same morphing attacks and morphing dataset across different quality measures used as MAD scores and the bona fide source used to fix the BPCER threshold. MagFace shows the lowest APCER for StyleGAN 2 generated morphs both in the FRLL-Morphs (APCER=0.0123), FERET-Morphs (APCER=0.0813), and the FRGC-Morphs (APCER=0.0861). In comparison to StyleGAN 2 attacks, the APCER produced by MagFace is higher for landmark-based attacks, such as FaceMorpher, OpenCV, and WebMorph. On the other hand, the inverted CNNIQA shows low APCER values on landmark-based attacks, both within the same morphing dataset or in other morphing datasets, this is more significant for the FRLL-Morphs and FERET-Morphs. 
Looking at the third row showing APCER results for FRGC-Morphs, the inverted dipIQ outperforms the inverted CNNIQA in terms of lower APCER for landmark-based morphing attacks. 

For detecting the morphing attacks in the LMA-DRD datasets, MagFace outperforms all other IQA and FIQA measures by a large margin when the BPCER threshold is set on the same bona fide source. In both cases of LMA-DRD (D) and LMA-DRD (PS) datasets, MagFace and the inverted rankIQ clearly outperform all other quality measures based on IQA. The reason could be the mitigated image artifacts after the re-digitization process rendering the FIQA methods more superior detection performance.

In the case of the MorGAN dataset, due to its relatively low image resolution, we focused only on the intra-class performance and its APCER. The lowest APCER is again observed for MagFace with 0.61 and 0.66 both for landmark-based OpenCV method and MorGAN-based morphing attacks respectively, which is still very high. 

% sections with ACER table 

\paragraph{Given an overall MAD performance measure, how well do quality measures perform as MADs across datasets and attacks?}

To answer that, we investigate the ACER values scored on different datasets when the BPCER is fixed on each bona fide source individually. 
Table \ref{tab:ACER} shows the ACER values, which is the balanced error between BPCER and APCER. It shows a clearer trade-off between bona fide and attack detection. A smaller ACER indicates the overall high performance of using this quality measure for the MAD task. This value is calculated when fixing the decision threshold such that the BPCER for the bona fide source shown in the first column is at 0.2. This threshold is then used to calculate the ACER, by adding the BPCER and APCER scores at this decision threshold and dividing by 2. 

The ACER values in Table \ref{tab:ACER} provide supporting findings to those extracted for the APCER in Table \ref{tab:APCER} and BPCER values in Table \ref{tab:BPCER}. The bold number shows the lowest ACER values determined when fixing the BPCER of the individual bona fide source in the first column at 0.2. MagFace as the detection metric performs more stable across all different kinds of morphing attacks and morphing datasets. For FRLL-Morphs, FERET-Morphs, and FRGC-Morphs, MagFace shows among other quality measures the best ACER for detecting morphing attacks such as StyleGAN 2 and also the attacks in LMA-DRD. The inverted CNN and dipIQ share the success for detecting landmark-based morphing attacks with minimum ACER value in the individual experimental settings. For the re-digitized LMA-DRD dataset, BRISQUE shows good error performance in terms of minimum ACER. 

The right-most column in Table \ref{tab:ACER} shows the \textit{mean} average ACER value across all morphing attacks and morphing datasets for each IQA and FIQA measure. We also marked all \textit{mean} ACER values below 0.3 in bold for better visibility. 
Based on this averaged ACER value, we observe that several quality/utility measures show stable ACER in the MAD task across different morphing attacks and morphing datasets. Mostly MagFace, but also the inverted CNNIQA, and the inverted dipIQ all perform equally well under specific experimental conditions. While FIQA measures work well on StyleGAN 2 generated morphing attacks, IQA measures work well in detecting facial landmark-based morphing attacks. The good performance of these measures is reflected in the low ACER values. Thus the quality measures of MagFace and CNNIQA can be used as unsupervised MAD with the overall performance of ACER below 0.30 under the diverse and unseen bona fide sources and attack variations. 
This relatively high MAD performance by MagFace and CNNIQA might be explained by their design concept. The MagFace quality is the magnitude of the sample embedding using the network trained as proposed in \cite{meng2021magface}. This is based on training the FR model using a loss that adapts the penalty margin loss based on this magnitude, and thus links the closeness of a sample to its class center to the unnormalized embedding magnitude. A morphed face image is designed to be close to multiple classes (identities), and thus is typically not expected to be extremely close to an individual class. Regarding the CNNIQA quality is based on processing image patches. This property is proven in \cite{kang2014convolutional} to be extremely sensitive to local distortions which makes it a perfect choice for local quality estimation. This corresponds to being sensitive to the local artifacts caused by the different morphing processes as discussed and shown in \cite{ReGenMorph}. One can take advantage of the different properties of image IQA and FIQA methods to build a possibly stronger MAD based on quality measures. This can be performed by quality score-level information fusion approach \cite{DBLP:conf/icpram/DamerOS13}, which, after score normalization \cite{DBLP:conf/eusipco/DamerBT0K18}, can give relative weights that correspond to the estimated performance of each of the fused methods \cite{DBLP:conf/eusipco/DamerON14,DBLP:conf/fusion/DamerON14}.

To sum up, our investigation on the detectability of morphing attacks using the quality measures shows that FIQA and IQA metrics can be used as unsupervised attack detection measures under different attack scenarios. While MagFace from FIQA performs relatively better on GAN-based morphing attacks, the inverse CNNIQA from IQA measure performs relatively better on landmark-based morphing attacks.

\section{Take-home-messages}
\label{sec:take-home-messages}
In this section, we summarize and highlight the most important outcomes discussed in details in Section \ref{ch:discussion}. 

\begin{itemize}
    \item All face morphing techniques do introduce a change in FIQ and IQ measures of face images in comparison to their corresponding bona fide samples. Such a shift was larger and more consistent in some FIQ/IQ methods than in others, e.g. MagFace, CNNIQA, and dipIQ showed relatively stronger quality score shifts. This shift directly corresponds to a quality separability between bona fide and attacks.
    \item The proposed use of FIQ and IQ measures as unsupervised MAD methods produce a stable and generalizable MAD performance for some of the quality measures. We noticed that setting an MAD decision threshold can produce relatively generalizable MAD decisions across bona fide sources and morphing techniques. This is especially the case for MagFace, CNNIQA, and dipIQ, as expected given the stable effect of different morphing techniques on their quality scores.
    \item Following the above-mentioned observations, we showed that using quality measures, such as MagFace and CNNIQA, as unsupervised MADs results in ACER values below 30\% on a diverse set of unknown data sources and morphing techniques. 
\end{itemize}

\section{Conclusion}
\label{ch:conclusion}

Given the complexity of MAD and the threat the morphing operation posed on the automatic face recognition system, we studied the effect of morphing on image quality and utility. We found a general quality shift in morphed images compared to bona fide images by using these investigated IQA and FIQA measures. However, these quality measures behave differently on landmark-based and GAN-based morphing attacks, yet to different degrees. 

Based on the observed separability in the quality measures between the bona fide and morph samples, we theorize that such measures can be used as unsupervised MADs, where the quality score can act as an MAD detection score.
Our analyses were successful in showing that using these quality measures to differentiate between morphing attacks and bona fide samples, quality measures such as MagFace and inverted CNNIQA can consistently lead to ACER values below 0.30 on completely unknown bona fide sources and attack methods.
Based on that, we conclude that these quality measures represent useful decision measures for practitioners designing applications for the MAD task as a stand-alone MAD indicator, or in future work, as a supporting measure for MAD decisions.

\section*{Acknowledgment}
This research work has been funded by the German Federal Ministry of Education and Research and the Hessian Ministry of Higher Education, Research, Science and the Arts within their joint support of the National Research Center for Applied Cybersecurity ATHENE.

\ifCLASSOPTIONcaptionsoff
  \newpage
\fi



%\bibliographystyle{IEEEtran}

\bibliographystyle{ieee_fullname}
\bibliography{egbib}



%\vfill


\end{document}